\documentclass{article}

\usepackage[final]{corl_2022} 
\usepackage{booktabs}       
\usepackage{nicefrac}       

\usepackage[numbers]{natbib}
\usepackage{multicol}
\usepackage{xcolor}         
\usepackage{amsmath, amsfonts}
\usepackage{pifont}
\usepackage{amsthm}
\usepackage[ruled,vlined,linesnumbered]{algorithm2e}
\newtheorem{condition}{Condition}
\newtheorem{theorem}{Theorem}
\newtheorem{lemma}{Lemma}

\newtheorem{assumption}{Assumption}
\usepackage{hhline}
\usepackage{graphicx}
\usepackage{floatrow}
\usepackage{sidecap}
\usepackage{wrapfig}
\usepackage[figuresleft]{rotating}

\usepackage{amssymb}
\usepackage{pifont}
\newcommand{\cmark}{\ding{51}}%
\newcommand{\xmark}{\ding{55}}%

\usepackage{enumitem}

\title{Temporal Logic Imitation: Learning Plan-Satisficing
Motion Policies from Demonstrations}

\author{
  Yanwei Wang\\
  MIT\\
  \And
  Nadia Figueroa\\
  University of Pennsylvania\\
  \And
  Shen Li \\
  MIT\\
  \And
  Ankit Shah \\
  Brown University\\
  \And
  Julie Shah \\
  MIT\\
}

\begin{document}
\maketitle
\vspace{-20pt}
\begin{abstract}
Learning from demonstration (LfD) has succeeded in tasks featuring a long time horizon. However, when the problem complexity also includes human-in-the-loop perturbations, state-of-the-art approaches do not guarantee the successful reproduction of a task. In this work, we identify the roots of this challenge as the failure of a learned continuous policy to satisfy the discrete plan implicit in the demonstration. By utilizing modes (rather than subgoals) as the discrete abstraction and motion policies with both mode invariance and goal reachability properties, we prove our learned continuous policy can simulate any discrete plan specified by a linear temporal logic (LTL) formula. Consequently, an imitator is robust to both task- and motion-level perturbations and guaranteed to achieve task success. \textbf{Project page: }\url{https://yanweiw.github.io/tli/}
\end{abstract}
\vspace{-8.5pt}

\begin{figure}[!h] 
      \includegraphics[width=0.25\textwidth]{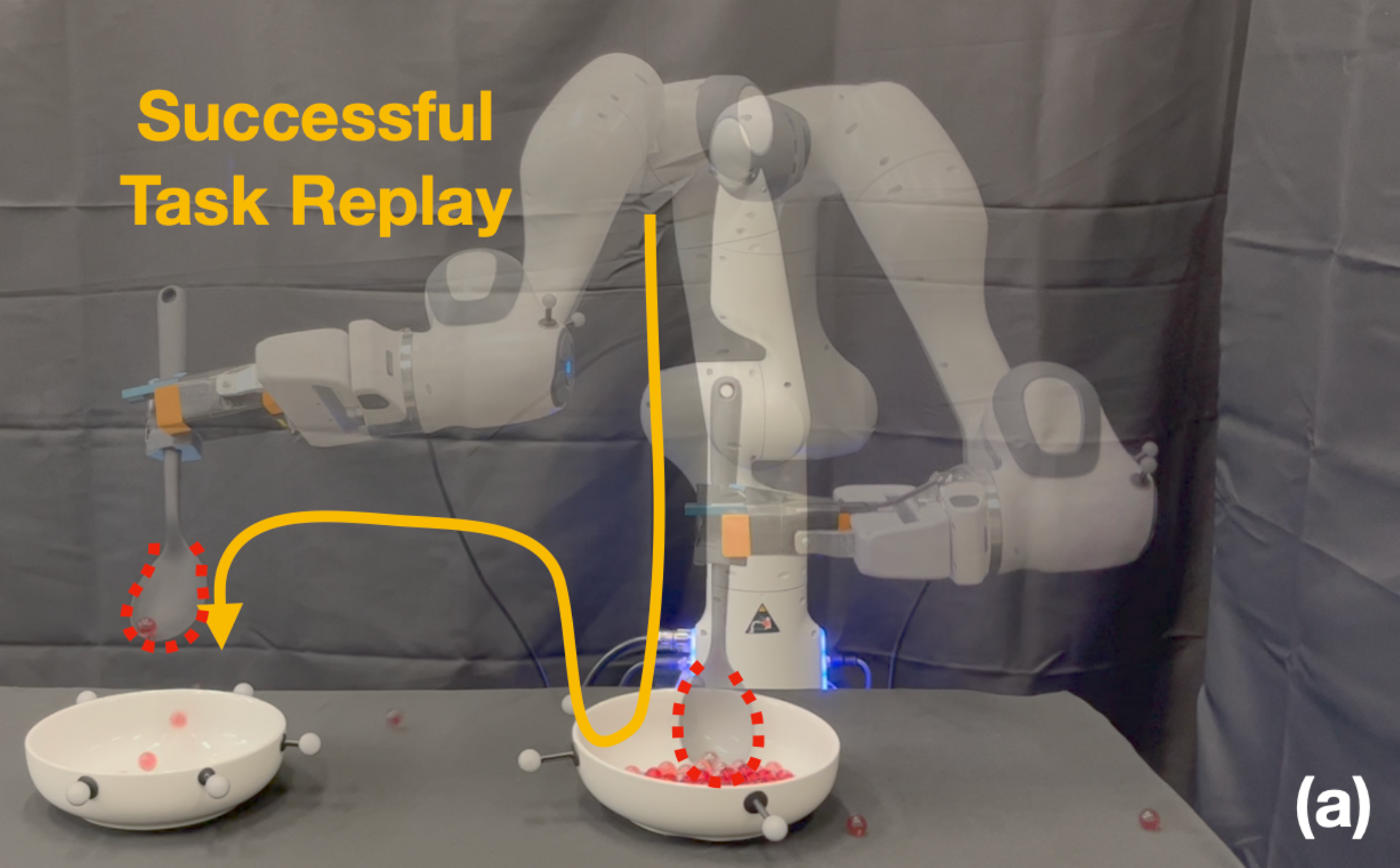}\includegraphics[width=0.25\textwidth]{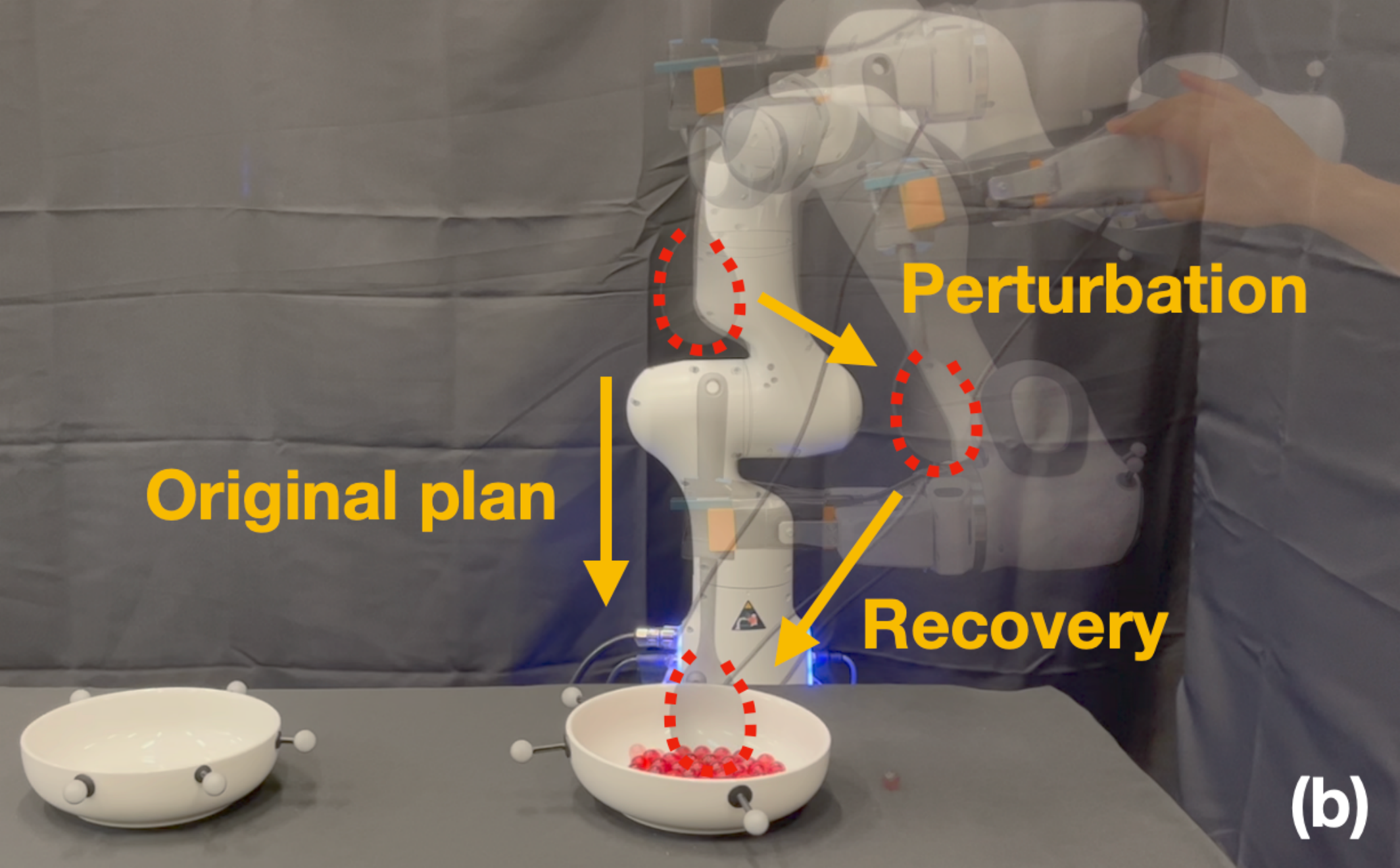}\includegraphics[width=0.25\textwidth]{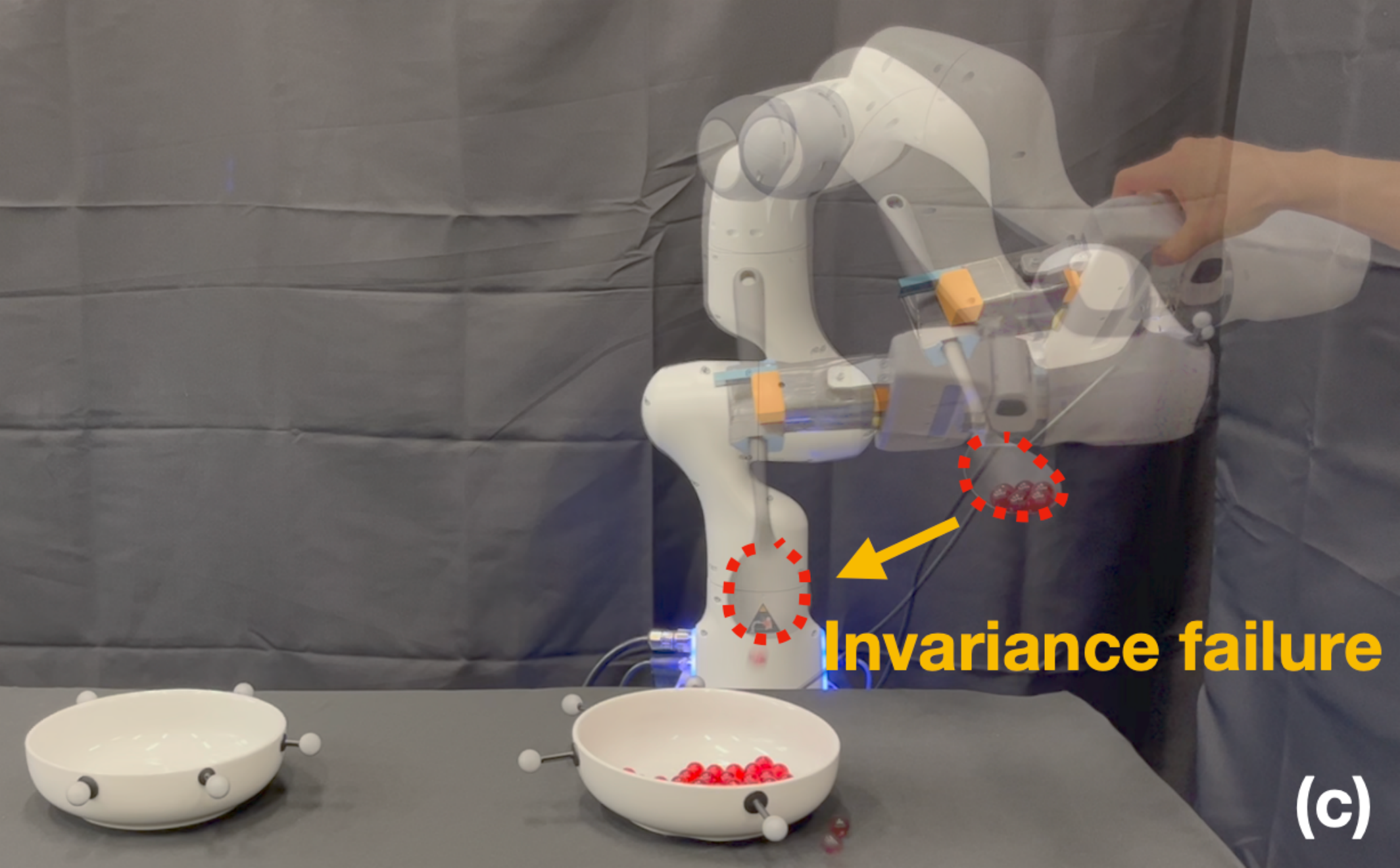}\includegraphics[width=0.25\textwidth]{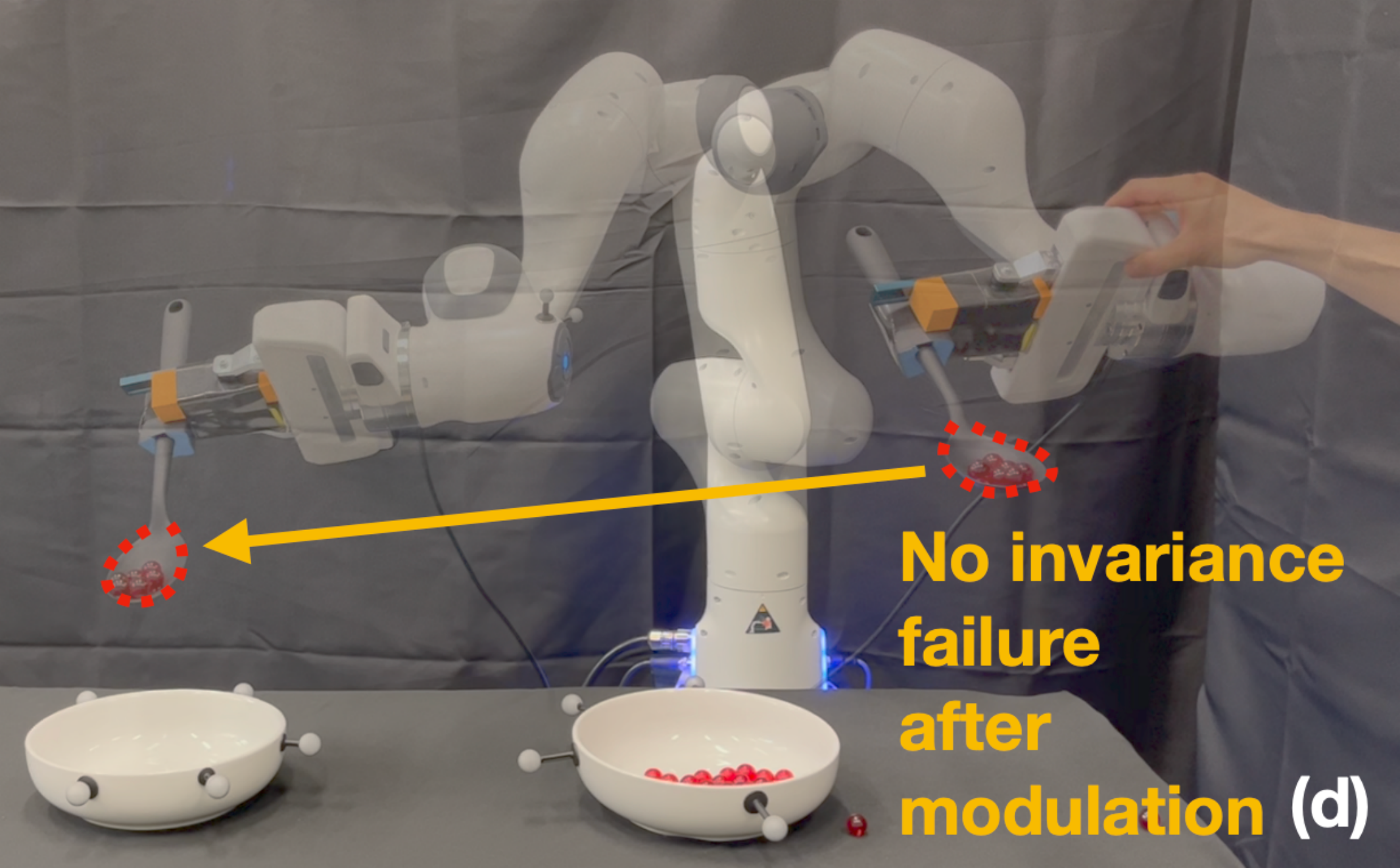}
     
      \caption{\footnotesize \textbf{(a)} A successful replay of the scooping task. The robot \textbf{(b)} is robust to motion-level perturbations; \textbf{(c)} experiences an invariance failure (i.e., drops material) after a task-level perturbation; and \textbf{(d)} re-scoops after a task-level perturbation, avoiding failure after DS motion policy modulation. \label{fig:robot_intro}}
\end{figure}

\vspace{-11pt}
\section{Introduction}
\vspace{-7.5pt}
In prior work, learning from demonstration (LfD) \cite{argall2009survey, ravichandar2020recent} has successfully enabled robots to accomplish multi-step tasks by segmenting demonstrations (primarily of robot end-effector or tool trajectories) into sub-tasks/goals \cite{ekvall2008robot, grollman2010incremental, medina2017learning, gupta2019relay, mandlekar2020learning, pirk2020modeling}, phases \cite{pastor2012towards, kroemer2015towards}, keyframes \cite{akgun2012trajectories, perez2017c}, or skills/primitives/options \cite{konidaris2012robot, niekum2013incremental, fox2017multi, figueroa2019high}. Most of these 
abstractions assume reaching subgoals sequentially will deliver the desired outcomes; however, successful imitation of many manipulation tasks with spatial/temporal constraints cannot be reduced to imitation at the motion level unless the learned motion policy also satisfies these constraints. This becomes highly relevant if we want robots to not only imitate but also generalize, adapt and be robust to perturbations imposed by humans, who are in the loop of task learning and execution. LfD techniques that learn stable motion policies with convergence guarantees (e.g., Dynamic Movement Primitives (DMP) \cite{saveriano2021dmp}, Dynamical Systems (DS) \cite{DSbook}) are capable of providing such desired properties but only at the motion level. As shown in Fig.~\ref{fig:robot_intro} (a-b) a robot can successfully replay a soup-scooping task while being robust to physical perturbations with a learned DS. Nevertheless, if the spoon orientation is perturbed to a state where all material is dropped, as seen in Fig.~\ref{fig:robot_intro} (c), the motion policy will still lead the robot to the target, unaware of the task-level failure or how to recover from it. To alleviate this, we introduce an imitation learning approach that is capable of i) reacting to such task-level failures with Linear Temporal Logic (LTL) specifications, and ii) modulating the learned DS motion policies to avoid repeating those failures as shown in Fig.~\ref{fig:robot_intro} (d).

\begin{figure}[!h]
    \vspace{-25pt}
    \includegraphics[trim={0cm 0cm 0 0},width=\textwidth, clip]{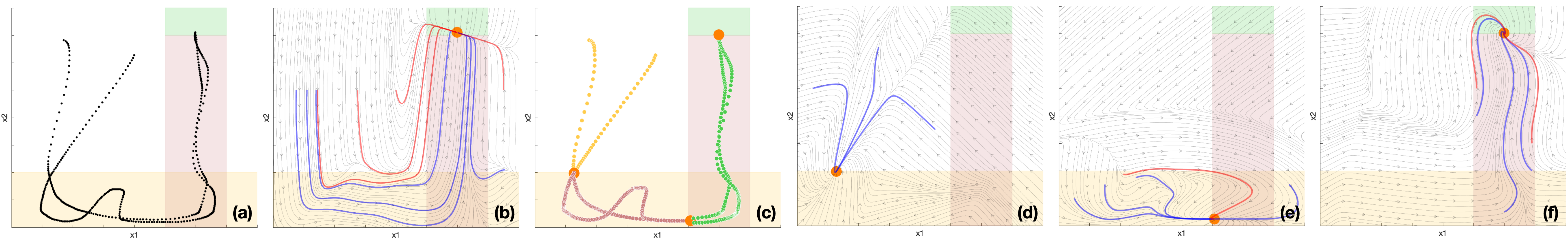}
    \caption{\footnotesize Mode abstraction of a 2D soup-scooping task: $x_1$ and $x_2$ denote the spoon's orientation and distance to the soup. \textbf{(a)} Task: To move the spoon's configuration from the white region (spoon without soup) $\Rightarrow$ yellow region (spoon in contact with soup) $\Rightarrow$ pink region (spoon holding soup) $\Rightarrow$ green region (soup at target). (Note that transitions (white $\Rightarrow$ pink) and (white $\Rightarrow$ green) are not physically realizable.) Black curves denote successful demonstrations. \textbf{(b)} Learning DS policies \cite{figueroa2018physically} over unsegmented data can result in successful task replay (blue trajectories), but lacks a guarantee due to invalid transitions (red trajectories). \textbf{(c)} Trajectories are segmented into three colored regions (modes) with orange attractors. \textbf{(d-f)} Learning DSs on segments may still result in \textit{invariance failures} (i.e., traveling outside of modes as depicted by red trajectories).\label{fig:intro} \vspace{-15pt}}
\end{figure}
\textbf{Example} We demonstrate that successfully reaching a goal via pure motion-level imitation does not imply successful task execution. The illustrations in Fig. \ref{fig:intro} represent a 2D simplification of the soup-scooping task, where task success requires a continuous trajectory to simulate a discrete plan of consecutive transitions through the colored regions. Human demonstrations, shown in Fig.~\ref{fig:intro}~\textbf{(a)}, are employed to learn a DS policy \cite{figueroa2018physically}, depicted by the streamlines in Fig.~\ref{fig:intro}~\textbf{(b)}. The policy is stress-tested by applying external perturbations, displacing the starting states of the policy rollouts. As shown, only blue trajectories succeed in the task, while the red ones fail due to discrete transitions that are not physically realizable (e.g., white $\Rightarrow$ pink). As shown in Fig.~\ref{fig:intro}~ \textbf{(c-f)}, even if the demonstrations are further segmented by subgoals (and corresponding DS policies are learned), this issue is not mitigated. While one could treat this problem as covariate shift and solve it by asking a human for more demonstrations \cite{ross2011reduction}, in this work, we frame it as the mismatch between a learned continuous policy and a discrete task plan specified by the human in terms of a logical formula. Specifically, the core challenges illustrated by this example are two-fold: 1) subgoals only impose point constraints that are insufficient to represent the boundary of a discrete abstraction; and 2) the continuous policy can deviate from a demonstrated discrete plan when perturbed to unseen parts of the state space, and is incapable of replanning to ensure all discrete transitions are valid. 

To address these challenges, our proposed approach employs ``modes” as discrete abstractions. We define a \textit{mode} as a set of robot and environment configurations that share the same sensor reading \cite{van2000introduction, garrett2021integrated}; e.g., in Fig. \ref{fig:intro}, each colored region is a unique mode, and every mode has a boundary that imposes path constraints on motion policies. Additionally, we use a task automaton as a receding-horizon controller that replans when a perturbation causes the system to travel outside a mode boundary and triggers an unexpected sensor change; e.g., detecting a transition from yellow $\Rightarrow$ white instead of the desired yellow $\Rightarrow$ pink will result in a new plan: white $\Rightarrow$ yellow $\Rightarrow$ pink $\Rightarrow$ green. In this work, we synthesize a task automaton from a linear temporal logic formula (LTL) that specifies all valid mode transitions. We denote the problem of learning a policy that respects these mode transitions from demonstrations as \textit{temporal logic imitation} (TLI). In contrast to temporal logic planning (TLP) \cite{kress2018synthesis}, where the workspace is partitioned into connected convex cells with known boundaries, we do not know the precise mode boundaries. Consequently, the learned policy might prematurely exit the same mode repeatedly, causing the task automaton to loop without termination. To ensure any discrete plan generated by the automaton is feasible for the continuous policy, the bisimulation criteria \cite{alur2000discrete, fainekos2005temporal} must hold for the policy associated with each mode. Specifically, any continuous motion starting in any mode should stay in the same mode \textbf{(invariance)} until eventually reaching the next mode \textbf{(reachability)}. The violations of these conditions are referred to as \textit{invariance failures} and \textit{reachability failures} respectively.

\textbf{Contributions} First, we investigate TLP in the setting of LfD and introduce TLI as a novel formulation to address covariate shift by proposing imitation with respect to a mode sequence instead of a motion sequence. Second, leveraging modes as the discrete abstraction, we prove that a state-based continuous behavior cloning (BC) policy with a global stability guarantee can be modulated to simulate any LTL-satisficing discrete plan. Third, we demonstrate that our approach LTL-DS, adapts to task-level perturbations via an LTL-satisficing automaton's replanning and recovers from motion-level perturbations via DS' stability during a multi-step, non-prehensile manipulation task.

\vspace{-10pt}
\section{Related Works} 
\vspace{-8.5pt}
\label{sec:related}
\textbf{Temporal Logic Motion Planning} LTL is a task specification language widely used in robot motion planning \cite{belta2007symbolic, wolff2013automaton, plaku2016motion, kress2018synthesis}. Its ease of use and efficient conversion \cite{piterman2006synthesis} to an automaton have spurred substantial research into TLP \cite{fainekos2005temporal, kress2009temporal, decastro2015synthesis}, which studies how to plan a continuous trajectory that satisfies an LTL formula. However, TLP typically assumes known workspace partitioning and boundaries \textit{a priori}, both of which are unknown in the rarely explored TLI setting. While a robot can still plan in uncertain environments \cite{ayala2013temporal, lahijanian2016iterative}, LfD bypasses the expensive search in high-dimensional space. Recent works \cite{innes2020elaborating, puranic2021learning} have considered temporal logic formulas as side-information to demonstrations, but these formulas are treated as additional loss terms or rewards and are not guaranteed to be satisfied. The key motivation for using LTL is to generate a reactive discrete plan, which can also be achieved by a finite state machine \cite{niekum2013incremental} or behavior tree \cite{li2021reactive}.\\
\noindent \textbf{Behavior Cloning} We consider a subclass of LfD methods called state-based behavior cloning (BC) that learns the state-action distribution observed during demonstrations \cite{osa2018algorithmic}. DAGGER \cite{ross2011reduction}, a BC-variant fixing covariate shift, could reduce the invariance failures depicted in Fig. \ref{fig:intro}, but requires online data collection, which our framework avoids with an LTL specification. To satisfy goal reachability, we employ a DS-based LfD technique \cite{khansari2011learning}. Alternatives to this choice include certified NN-based methods \cite{neumann2013neural, dawson2022safe}, DMPs \cite{ijspeert2013dynamical}, partially contracting DS \cite{ravichandar17a}, and Euclideanizing-flows \cite{rana2020learning}. To satisfy mode invariance, we modulate the learned DS to avoid invariance failure as state-space boundaries \cite{khansari2012dynamical}, similar to how barrier functions are learned to bound a controller \cite{robey2020learning, saveriano2019learning, dawson2022survey}.
\noindent \textbf{Multi-Step Manipulation} Prior LfD works \cite{konidaris2012robot, niekum2013incremental, kroemer2015towards, ye2017guided} tackle multi-step manipulation by segmenting demonstrations via a hidden Markov model. Using segmented motion trajectories, \cite{konidaris2012robot} learned a skill tree, \cite{niekum2013incremental} learned DMPs, \cite{kroemer2015towards} learned phase transitions, and \cite{bowen2014closed} learned a task model. Most of these works assume a linear sequence of prehensile subtasks (pick-and-place) without considering how to replan when unexpected mode transitions happen. \cite{ye2017guided, bowen2014closed} considered a non-prehensile scooping task similar to ours, but their reactivity only concerned collision avoidance in a single mode. \cite{rajeswaran2017learning, gupta2019relay} improved BC policies with RL, but offered no guarantee of task success.
\vspace{-10pt}

\section{Temporal Logic Imitation: Problem Formulation}
\label{sec:tli}
\vspace{-8.5pt}
Let $x \in \mathbb{R}^n$ represent the $n$-dimensional continuous state of a robotic system; e.g., the robot's end-effector state in this work. Let $\alpha = [\alpha_1,...,\alpha_m]^T \in \{0, 1\}^m$ be an $m$-dimensional discrete sensor state that uniquely identifies a mode $\sigma = \mathcal{L}(\alpha)$. We define a system state as a tuple, $s = (x, \alpha) \in \mathbb{R}^n \times \{0, 1\}^m$. Overloading the notation, we use $\sigma \in \Sigma$, where $\Sigma = \{\sigma_i\}_{i=1}^{\mathcal{M}}$, to represent the set of all system states within the same mode---i.e., $ \sigma_i = \{s=(x, \alpha) \mid \mathcal{L}(\alpha) = \sigma_i\}$. In contrast, we use $\delta_i=\{x | s=(x,\alpha) \in \sigma_i\}$ to represent the corresponding set of robot states. Note $x$ cannot be one-to-one mapped to $s$, e.g., a level spoon can be either empty or holding soup. Each mode is associated with a goal-oriented policy, with goal $x^*_i\in\mathbb{R}^n$. A successful policy that accomplishes a multi-step task $\tau$ with a corresponding LTL specification $\phi$ can be written in the form:
\vspace{-0.15pt}
\begin{equation}
\label{eq:imitation_pi}
\begin{aligned}
& \dot{x} =  \pi(x,\alpha; \phi) = \Sigma_{i=1}^{\mathcal{M}}\delta_{\Omega_{\phi}(\alpha)\sigma_i} f_i(x;\theta_i, x^*_i) 
\end{aligned}
\end{equation}
\vspace{-0.15pt}
with $\delta_{\Omega_{\phi}(\alpha)\sigma_i}$ being the Kronecker delta that activates a mode policy $f_i(x;\theta_i,x^*_i):\mathbb{R}^{n}\rightarrow\mathbb{R}^n$ encoded by learnable parameters $\theta_i$ and goal $x^*_i$. Mode activation is guided by an LTL-equivalent automaton $\Omega_{\phi}(\alpha) \rightarrow \sigma_i$ choosing the next mode $\sigma_i$ based on the current sensor reading $\alpha$. \\
\textbf{Demonstrations} Let demonstrations for a task $\tau$ be $\Xi= \{\{x^{t,d},\dot{x}^{t,d}, \alpha^{t,d}\}_{t=1}^{T_d}\}_{d=1}^{D}$ where $x^{t,d}\text{, } \dot{x}^{t,d}\text{, } \alpha^{t,d}$ are robot state, velocity, and sensor state at time $t$ in demonstration $d$, respectively, and $T_d$ is the length of each $d$-th trajectory. A demonstration is successful if the continuous motion traces through a sequence of discrete modes that satisfies the corresponding LTL task specification. \\
\textbf{Perturbations} External perturbations, which many works in Sec. \ref{sec:related} avoid, constitute an integral part of our task complexity. Specifically, we consider (1) motion-level perturbations that displace a continuous motion within the same mode, and (2) task-level perturbations that drive the robot outside of the current mode. Critically, motion-level perturbations do not cause a plan change instantaneously, but they can lead to future unwanted mode transitions due to covariate shift. \\
\textbf{Problem Statement} Given (1) an LTL formula $\phi$ specifying valid mode transitions for a task $\tau$, (2) 
sensors that detect each mode abstraction defined in $\phi$, and (3) successful demonstrations $\Xi$, we seek to learn a policy defined in Eq. \ref{eq:imitation_pi} that generates continuous trajectories guaranteed to satisfy the LTL specification despite arbitrary external perturbations. 
\vspace{-8.5pt}

\section{Preliminaries} 
\vspace{-8.5pt}
\label{sec:prelims}
\subsection{LTL Task Specification}\label{sec:ltl_spec} 
\vspace{-8.5pt}
LTL formulas consist of atomic propositions (AP), logical operators, and temporal operators \cite{shah2018bayesian, kress2018synthesis}. Let $\Pi$ be a set of Boolean variables; an infinite sequence of truth assignments to all APs in $\Pi$ is called the trace $[\Pi]$. The notation $[\Pi],t\models\phi$ means the truth assignment at time $t$ satisfies the LTL formula $\phi$. Given $\Pi$, the minimal syntax of LTL can be described as:
\begin{equation} \label{eq:ltl}
    \phi ::= p \mid \neg\phi_1 \mid \phi_1 \vee \phi_2 \mid \textbf{X}\phi_1 \mid \phi_1\textbf{U}\phi_2
\end{equation}
where $p$ is any AP in $\Pi$, and $\phi_1$ and $\phi_2$ are valid LTL formulas constructed from $p$ using Eq. \ref{eq:ltl}. The operator \textbf{X} is read as `next,' and $\textbf{X}\phi_1$ intuitively means the truth assignment to APs at the next time step sets $\phi_1$ as true. \textbf{U} is read as `until' and, intuitively, $\phi_1\textbf{U}\phi_2$ means the truth assignment to APs sets $\phi_1$ as true until $\phi_2$ becomes true. Additionally, first-order logic operators $\neg$ (not), $\land$ (and), $\lor$ (or), and $\rightarrow$ (implies), as well as higher-order temporal operators \textbf{F} (eventually), and \textbf{G} (globally), are incorporated. Intuitively, $\textbf{F}\phi_1$ means the truth assignment to APs eventually renders $\phi_1$ true and $\textbf{G}\phi_1$ means truth assignment to APs renders $\phi_1$ always true from this time step onward. 
\vspace{-8.5pt}
\subsection{Task-Level Reactivity in LTL} \label{sec:fsm}
\vspace{-6.5pt}
To capture the reactive nature of a system given sensor measurements, the \textit{generalized reactivity (1)} (GR(1)) fragment of LTL \cite{piterman2006synthesis, kress2009temporal} can be used. Let the set of all APs be $\Pi = \mathcal{X}\cup\mathcal{Y}$, where sensor states form environment APs $\mathcal{X} = \{\alpha_1,...,\alpha_m\}$ and mode symbols form system APs $\mathcal{Y} = \{\sigma_1,...,\sigma_l\}$. A GR(1) formula is of the form $\phi = (\phi_e \rightarrow \phi_s)$ \cite{piterman2006synthesis}, where $\phi_e$ models the assumed environment behavior and $\phi_s$ models the desired system behavior. Specifically,
\begin{equation} \label{eq:gr1}
    \phi_e=\phi^e_i\land \phi^e_t\land \phi^e_g, \quad\quad \phi_s=\phi^s_i\land \phi^s_t\land \phi^s_g
\end{equation}
$\phi^e_i$ and $\phi^s_i$ are non-temporal Boolean formulas that constrain the initial truth assignments of $\mathcal{X}$ and $\mathcal{Y}$ (e.g., the starting mode). $\phi^s_t$ and $\phi^e_t$ are LTL formulas categorized as safety specifications that describe how the system and environment should always behave (e.g., valid mode transitions). $\phi_g^s$ and $\phi_g^e$ are LTL formulas categorized as liveness specifications that describe what goal the system and environment should eventually achieve (e.g., task completion) \cite{kress2018synthesis}.
The formula $\phi$ guarantees the desired system behavior specified by $\phi_s$ if the environment is \textit{admissible}---i.e., $\phi_e$ is true---and can be converted to an automaton $\Omega_{\phi}$ that plans a mode sequence satisfying $\phi$ by construction \cite{kress2009temporal}.
\vspace{-7.5pt}

\subsection{Motion-Level Reactivity in DS} \label{sec:ds_learning}
\vspace{-5.5pt}
Dynamical System \cite{figueroa2018physically} is a state-based BC method with a goal-reaching guarantee despite arbitrary perturbations. A DS policy can be learned from as few as a single demonstration and has the form:

\vspace{-10pt}

\begin{minipage}{.4\textwidth}
\begin{equation}
\label{eq:ds_eq}
  \dot{x} = f(x) = \sum_{k=1}^{K}\gamma_k(x)(A^kx+b^k)
\end{equation}
\end{minipage}\hspace{1pt} 
\begin{minipage}{.575\textwidth}
\begin{equation}
\label{eq:stability}
  \begin{cases}
    (A^k)^T P+PA^k =Q^k, Q^k=(Q^k)^T \prec 0 \\
    b^k=-A^k x^*
    \end{cases}
    \forall k
\end{equation}
\end{minipage}

where $A^k\in \mathbb{R}^{n\times n}$, $b^k\in \mathbb{R}^n$ are the k-th linear system parameters, and $\gamma_k(x):\mathbb{R}^n \rightarrow \mathbb{R}^+$ is the mixing function. To certify global asymptotic stability (G.A.S.) of Eq. \ref{eq:ds_eq}, a Lyapunov function $V(x)=(x-x^*)^T P(x-x^*)$ with $P=P^T\succ 0$, is used to derive the stability constraints in Eq. \ref{eq:stability}. Minimizing the fitting error of Eq. \ref{eq:ds_eq} with respect to demonstrations $\Xi$ subject to constraints in Eq. \ref{eq:stability} yields a non-linear DS with a stability guarantee \cite{figueroa2018physically}. To learn the optimal number $K$ and mixing function $\gamma_k(x)$ we use the Bayesian non-parametric GMM fitting approach presented in \cite{figueroa2018physically}. 

\vspace{-7.5pt}

\subsection{Bisimulation between Discrete Plan and Continuous Policy}
\vspace{-7.5pt}
To certify a continuous policy will satisfy an LTL formula $\phi$, one can show the policy can simulate any LTL-satisficing discrete plan of mode sequence generated by $\Omega_{\phi}$. To that end, every mode's associated policy must satisfy the following bisimulation conditions \cite{fainekos2005temporal, kress2018synthesis}:
\begin{condition}[\textbf{Invariance}] Every continuous motion starting in a mode must remain within the same mode while following the current mode's policy; i.e., $\forall i \ \forall t \ (s^0 \in \sigma_i \rightarrow s^t \in \sigma_i)$
\end{condition}
\begin{condition}[\textbf{Reachability}] 
Every continuous motion starting in a mode must reach the next mode in the demonstration while following the current mode's policy; i.e., $\forall i \ \exists T \ (s^0 \in \sigma_i \rightarrow s^{T} \in \sigma_j)$
\end{condition}

\vspace{-8.5pt}

\section{LTL-DS: Methodology}
\label{sec:ltlds}
\vspace{-8.5pt}
To solve the TLI problem in Sec. \ref{sec:tli}, we introduce a mode-based imitation policy---LTL-DS:
\begin{equation}
\label{eq:ltlds_pi}
\dot{x} = \pi(x,\alpha; \phi) = \textcolor{blue}{\underbrace{\Sigma_{i=1}^{\mathcal{M}}\delta_{\Omega_{\phi}(\alpha)\sigma_i}}_{\text{offline learning}}} \textcolor{orange}{\underbrace{M_i\big(x;\Gamma_i(x), x^*_i\big)}_{\text{online learning}}}\textcolor{blue}{\underbrace{f_i(x;\theta_i, x^*_i)}_{\text{offline learning}}},
\end{equation}
During offline learning, we synthesize the automaton $\Omega_{\phi}$ from $\phi$ as outlined in Sec. \ref{sec:fsm} and learn DS policies $f_i$ from $\Xi$ according to Sec. \ref{sec:ds_learning}. While the choice of DS satisfies the reachability condition as explained later, nominal DS rollouts are not necessarily bounded within any region. Neither do we know mode boundaries in TLI. Therefore, an online learning phase is necessary, where for each mode policy $f_i$ we learn an implicit function, $\Gamma_i(x): \mathbb{R}^n \rightarrow \mathbb{R^+}$, that inner-approximates the mode boundary in the state-space of a robot $x\in\mathbb{R}^n$. With a learned $\Gamma_i(x)$ for each mode, we can construct a modulation matrix $M_i$ that ensures each modulated DS---$M_if_i$---is mode invariant.
\vspace{-8.5pt}

\begin{figure}[!tbp]
  \vspace{-3pt}
  \centering
  \includegraphics[width=0.95\textwidth]{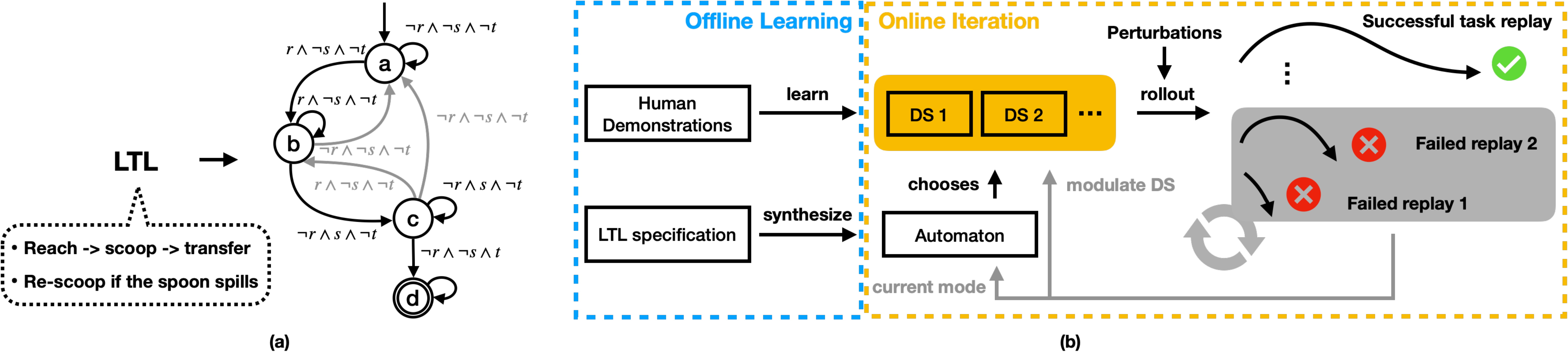}
  \caption{\footnotesize \textbf{(a)} Task automaton for a scooping task LTL. Mode $a, b, c, d$ are reaching, scooping, transporting, and done mode respectively. Atomic proposition $r, s, t$ denote sensing the spoon reaching the soup, soup on the spoon, and task success respectively. 
  During successful demonstrations, only mode transitions in black, $a \Rightarrow b \Rightarrow c \Rightarrow d$, are observed. Additional valid transitions in gray, $b \Rightarrow a$, $c \Rightarrow a$, and $c \Rightarrow b$, are given by the LTL to help recover from unexpected mode transitions. 
  \textbf{(b)} System flowchart of LTL-DS. 
\label{fig:method} \vspace{-10pt}}
\end{figure} 

\subsection{Offline Learning Phase}
\vspace{-7.5pt}
\textbf{Synthesis of LTL-Satisficing Automaton}
We convert an LTL to its equivalent automaton with \cite{duret.16.atva2}, which plans the next mode given the current sensor reading. Assuming all possible initial conditions for the system are specified in the LTL, the automaton is always deployed from a legal state.\\
\textbf{Sensor-based Motion Segmentation and Attractor Identification} Given demonstrations in $\Xi$ and accompanying sensor readings related to the set of $\mathcal{M}$ modes, we can automatically segment the trajectories into $\mathcal{M}$ clusters and corresponding attractor set $X^*$. Refer to Appendix \ref{sec:segmentation} for details.\\
\textbf{Ensuring Goal Reachability with Learned DS Mode Policies}
While any BC variant with a stability guarantee can satisfy reachability (see Sec. \ref{sec:related}), we focus on the G.A.S. DS formulation and learning approach defined in Section \ref{sec:ds_learning} that ensures every $x\in \mathbb{R}^n$ is guaranteed to reach $x^*_{i}$. By placing $x^*_{i}$ within the boundary set of $\delta_j$ for a mode $\sigma_j$, we ensure mode $\sigma_j$ is reachable from every $s$ in mode $\sigma_i$. Note $f(x)$ cannot model sensor dynamics in $\alpha$. Yet, we employ mode abstraction to reduce the imitation of a system state trajectory in $s$---which includes the evolution of both the robot and sensor state---to just a robot state trajectory in $x$. 
\vspace{-8.5pt}

\subsection{Online Learning Phase}
\label{sec:invariance}
\begin{figure}[!tbp]
    \vspace{-15pt}
    \includegraphics[trim={0cm 0cm 0 0},width=\textwidth, clip]{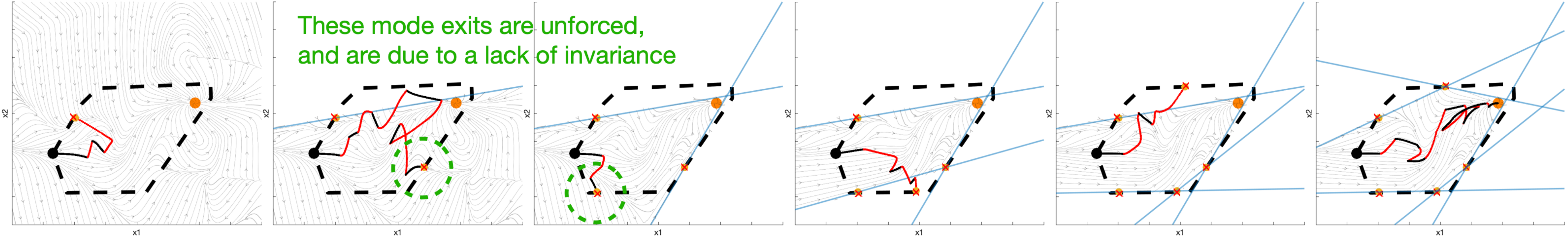}
    \caption{\footnotesize An illustration of iterative estimation of a mode boundary with cutting planes. A system enters a mode with an unknown boundary (dashed line) at the black circle, and is attracted to the goal at the orange circle. The trajectory in black shows the original policy rollout, and the trajectory in red is driven by perturbations. After the system exits the mode and before it eventually re-enters the same mode through replanning, a cut is placed at the last in-mode state (yellow circle) to bound the mode from the failure state (red cross). When the system is inside the cuts, it experiences modulated DS and never moves out of the cuts (flows moving into the cuts are not modulated); when the system is outside the cuts but inside the mode, it follows the nominal DS. Note only mode exits in black are invariance failures in need of modulation (green circles); mode exits in red are driven by perturbations to illustrate that more cuts lead to better boundary approximation.\label{fig:cut}\vspace{-15pt}}
\end{figure}

\vspace{-8.5pt}
\textbf{Iterative Mode Boundary Estimation via Invariance Failures} As shown in Fig. \ref{fig:intro}, DS can suffer from \textit{invariance} failures in regions without data coverage. Instead of querying humans for more data in those regions \cite{ross2011reduction}, we leverage sparse events of mode exits detected by sensors to estimate the unknown mode boundary. Specifically, for each invariance failure, we construct a cut that separates the failure state, $x^{T_f}$, from the mode-entry state, $x^{0}$, the last in-mode state, $x^{T_f-1}$, and the mode attractor, $x^*$. We ensure this separation constraint with a quadratically constrained quadratic program (QCQP) that searches for the normal direction (pointing away from the mode) of a hyper-plane that passes through each $x^{T_f-1}$ such that the plane's distance to $x^*$ is minimized. The intersection of half-spaces cut by hyper-planes inner approximates a convex mode boundary, as seen in Fig. \ref{fig:cut}. Adding cuts yields better boundary estimation, but is not necessary unless the original vector field flows out of the mode around those cuts. For more details, refer to Appendix \ref{sec:qcqp}.\\
\textbf{Ensuring Mode Invariance by Modulating DS} We treat each cut as a collision boundary that deflects DS flows following the approach in \cite{khansari2012dynamical,huber2019avoidance}. In our problem setting the mode boundary is analogous to a workspace enclosure rather than a task-space object. Let existing cuts form an implicit function, $\Gamma(x): \mathbb{R}^n \rightarrow \mathbb{R^+}$, where $\Gamma(x)<1$, $\Gamma(x)=1, \Gamma(x)>1$ denote the estimated interior, the boundary and the exterior of a mode. $0<\Gamma(x)<\infty$ monotonically increases as $x$ moves away from a reference point $x^r$ inside the mode. For $x$ outside the cuts, or inside but moving away from the cuts, we leave $f(x)$ unchanged; otherwise, we modulate $f(x)$ to not collide with any cuts as $\dot{x}=M(x)f(x)$ by constructing a modulation matrix $M(x)$ through eigenvalue decomposition:
\begin{equation}\label{eq:modulate}
    \begin{cases}
        M(x) = E(x)D(x)E(x)^{-1},~~ 
        E(x) = [\bold{r}(x)\text{ } \bold{e}_1(x)\text{ } ...\text{ } \bold{e}_{d-1}(x)],~~\bold{r}(x) = \frac{x-x^r}{\|x-x^r\|}\\ D(x) = 
        \textbf{diag(}\lambda_r(x), \lambda_{e_1}(x),...,\lambda_{e_{d-1}}(x)\textbf{)},~~
        \lambda_r(x) = 1-\Gamma(x),~~\lambda_e(x) = 1
    \end{cases}
\end{equation}
The full-rank basis $E(x)$ consists of a reference direction $\bold{r}(x)$ stemming from $x^r$ toward $x$, and $d-1$ directions spanning the hyperplane orthogonal to $\nabla\Gamma(x)$, which in this case is the closest cut to $x$. In other words, all directions $\bold{e}_1(x)...\bold{e}_{d-1}(x)$ are tangent to the closest cut, except $\bold{r}(x)$. By modulating only the diagonal component, $\lambda_r(x)$, with $\Gamma(x)$, we have $\lambda_r(x) \rightarrow 0$ as $x$ approaches the closest cut, effectively zeroing out the velocity penetrating the cut while preserving velocity tangent to the cut. Consequently, a modulated DS will not repeat invariance failures that its nominal counterpart experiences as long as the mode is bounded by cuts. Notice this modulation strategy is not limited to DS and can be applied to any state-based BC method to achieve mode invariance.
\vspace{-8.5pt}

\section{Proof}
\vspace{-7.5pt}
Next, we prove LTL-DS produces a continuous trajectory that satisfies an LTL specification. We start with assumptions and end with theorems. Detailed proofs are provided in Appendix \ref{sec:proof}. 
\begin{assumption} \label{eq:convexity}
    All modes are convex. 
\end{assumption}
\vspace{-2mm}
This assumption leads to the existence of at least one cut---i.e., the supporting plane \cite{boyd2004convex}, which can separate a failure state on the boundary from any other state within the mode. A corollary is that the boundary shared by two modes, which we call a guard surface, $G_{ij} = \delta_i \cap \delta_j$, is also convex. Since all transitions out of a mode observed during demonstrations reside on the mode boundary, their average location, which we use as the attractor for the mode, will also be on the boundary.

\begin{assumption} \label{eq:finiteness}
    There are a finite number of external perturbations of arbitrary magnitudes.
\end{assumption}
\vspace{-2mm}
Given zero perturbation, all BC methods should succeed in any task replay, as the policy rollout will always be in distribution. If there are infinitely many arbitrary perturbations, no BC methods will be able to reach a goal. In this work, we study the setting in between, where there are finitely many motion- and task-level perturbations causing unexpected mode exits. Environmental stochasticity is ignored, as its cumulative effects can also be simulated by external perturbations. 

\begin{assumption} \label{eq:non-oracle}
    Perturbations only cause transitions to modes already seen in the demonstrations.
\end{assumption}
\vspace{-2mm}
While demonstrations of all valid mode transitions are not required, they must minimally cover all possible modes. If a system encounters a completely new sensor state during online interaction, it is reasonable to assume that no BC methods could recover from the mode unless more information about the environment is provided. 
\begin{theorem} \label{eq:kc1}
    (Key Contribution 1) A nonlinear DS defined by Eq. \ref{eq:ds_eq}, learned from demonstrations, and modulated by cutting planes as described in Section \ref{sec:invariance} with the reference point $x^r$ set at the attractor $x^*$, will never penetrate the cuts and is G.A.S. at $x^*$. \hfill\textbf{Proof:} See Appendix \ref{sec:proof}.~~\qedsymbol{}
\end{theorem} 
\begin{theorem} (Key Contribution 2) \label{eq:kc2}
    The continuous trace of system states generated by LTL-DS defined in Eq. \ref{eq:ltlds_pi} satisfies any LTL specification $\phi$ under Asm. \ref{eq:convexity}, \ref{eq:finiteness}, and \ref{eq:non-oracle}.~\textbf{Proof:} See Appendix \ref{sec:proof}.\qedsymbol{} 
\end{theorem}
\vspace{-7.5pt}

\section{Experiments}
\vspace{-5.5pt}
\subsection{Single-Mode Invariance and Reachability}
\begin{figure}[!tbp]
\vspace{-15pt}
    \includegraphics[width=1\linewidth]{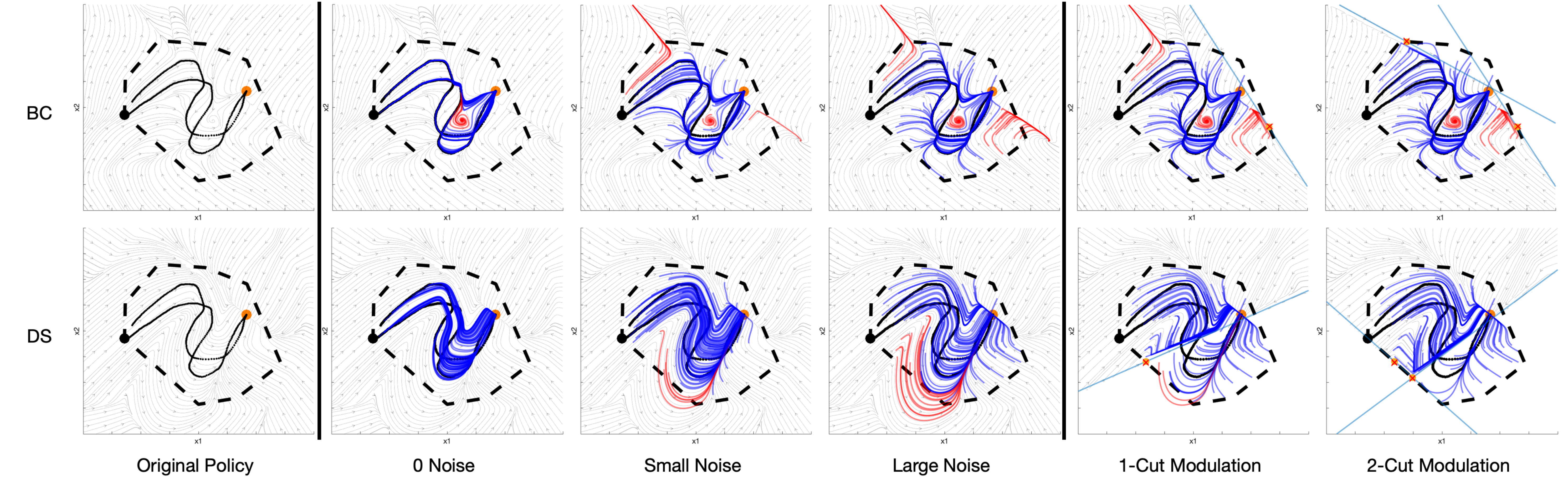}
\caption{\footnotesize Policy rollouts from different starting states for a randomly generated convex mode. The top row shows BC results, and the bottom row depicts DS results. The left column visualizes the nominal policies learned from two demonstrations (black trajectories) reaching the orange attractor. The middle columns add different levels of Gaussian noise to the initial states sampled from the demonstration distribution. Blue trajectories successfully reach the attractor, while red trajectories fail due to either invariance failures or reachability failures. (Note that these failures only occur at locations without data coverage.) The right columns show that cutting planes (blue lines) separate failures (red crosses) from last-visited in-mode states (yellow circles) and consequently bound both policies to be mode-invariant. Applying cutting planes to BC policies without a stability guarantee cannot correct reachability failures within the mode. More results in Appendix \ref{sec:single-mode}.}\label{fig:rollout}
\end{figure}

\begin{figure}[!tbp]
\vspace{-8pt}
\centering
\begin{minipage}{.51\textwidth}
    \resizebox{0.9\columnwidth}{!}{%
    \begin{tabular}{cccccc}
    \hline
    Policy & Reachability & Invariance & No Noise & Small Noise & Large Noise \\ \hline
    BC     & \color{red}\xmark & \color{red}\xmark  & 88.9 & 72.4 & 58.6  \\
    BC+mod & \color{red}\xmark & \color{green}{\cmark}  & 91.9 & 83.6 & 76.0  \\
    DS     & \color{green}{\cmark} & \color{red}\xmark  & 100  & 97.0 & 86.9  \\
    DS+mod & \color{green}{\cmark} & \color{green}{\cmark}  & \textbf{100}  & \textbf{100}  & \textbf{100}   \\ 
    \hline
    \end{tabular}}
    \vspace{3pt}\\
      \includegraphics[width=0.9\linewidth]{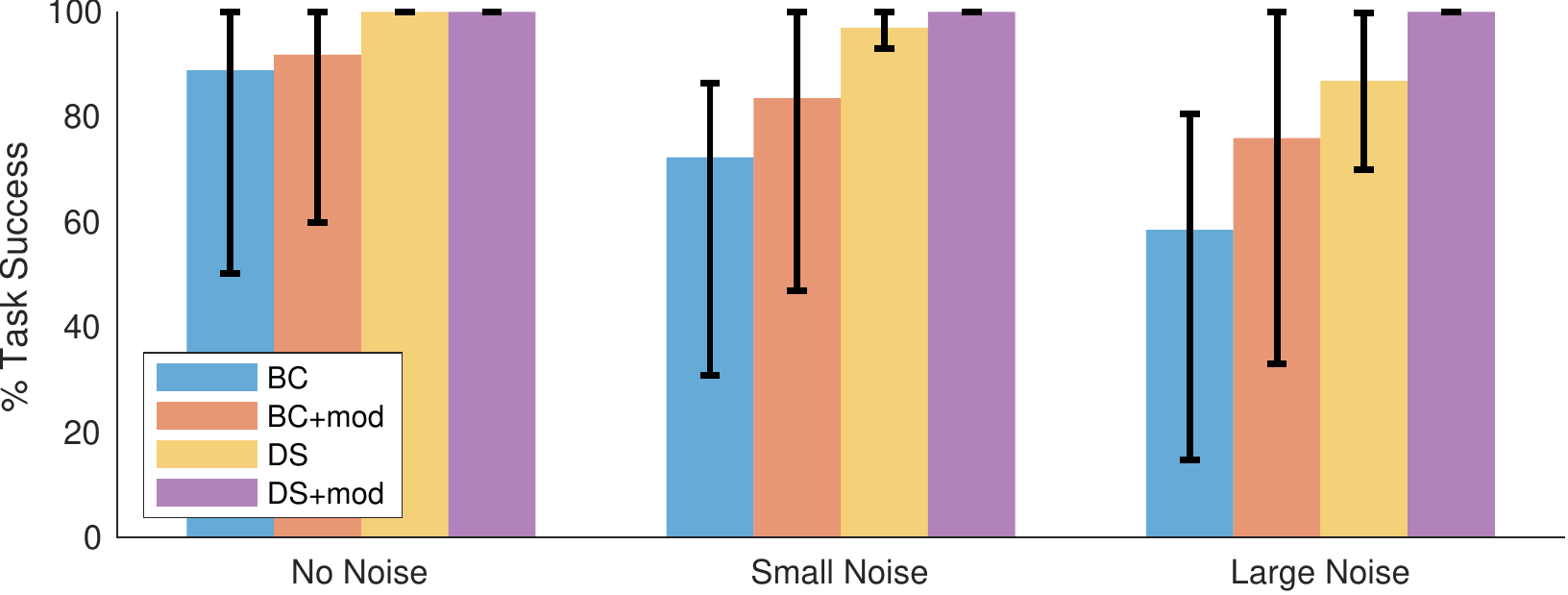}
\end{minipage}%
\begin{minipage}{.45\textwidth}
  \includegraphics[width=0.9\linewidth]{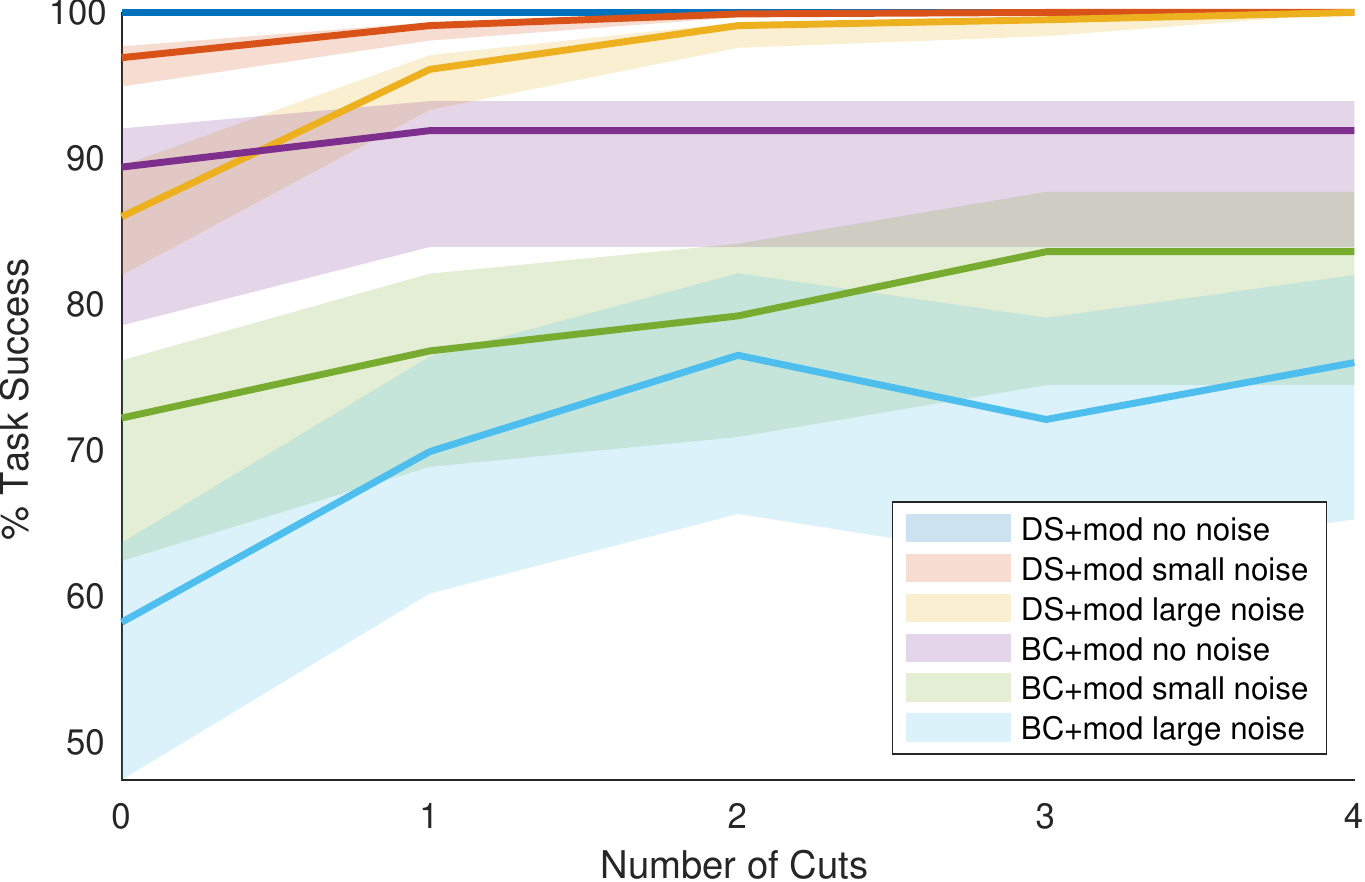}
  \caption{\footnotesize \textbf{(Left)} The success rate (\%) of a single-mode reaching task. As we begin to sample out of distribution by adding more noise to the demonstrated states, the BC's success rate degrades more rapidly than the DS'. After modulation, DS (+mod) maintains a success guarantee, which BC (+mod) falls short of due to the base policy's lack of a stability guarantee. \textbf{(Right)} Empirically, the invariance of a single mode requires only a finite number of cuts for a nominal policy with a stability guarantee. Regardless of the noise level, DS achieves a $100\%$ success rate after four cuts, while BC struggles to improve performance with additional cuts. Thick lines represent mean statistics and shaded regions show the interquartile range. More details in Appendix \ref{sec:single-mode}. \label{fig:result2} \vspace{-10.5pt}}
\end{minipage}
\end{figure}

\vspace{-7.5pt}
We show quantitatively  both reachability and invariance are necessary for task success. We compare DS and a NN-based BC policy (denoted as BC) to represent policies with and without a stability guarantee. Figure \ref{fig:rollout} shows that policy rollouts start to fail (turn red) as increasingly larger perturbations are applied to the starting states; however, DS only suffers from invariance failures, while BC suffers from both invariance and reachability failures (due to diverging flows and spurious attractors). Figure \ref{fig:rollout} (right) shows that all flows are bounded within the mode for both DS and BC after two cuts. In the case of DS, flows originally leaving the mode are now redirected to the attractor by the cuts; in the case of BC, while no flows leave the mode after modulation, spurious attractors are created, leading to reachability failures. This is a counterfactual illustration of Thm. \ref{eq:kc1}, that policies without a stability guarantee are not G.A.S. after modulation. Figure \ref{fig:result2} verifies this claim quantitatively and we empirically demonstrate that a stable policy requires only four modulation cuts to achieve a perfect success rate---which an unstable policy cannot be modulated to achieve.
\begin{figure}[!tbp]
\vspace{-1pt}
    \includegraphics[width=1\linewidth]{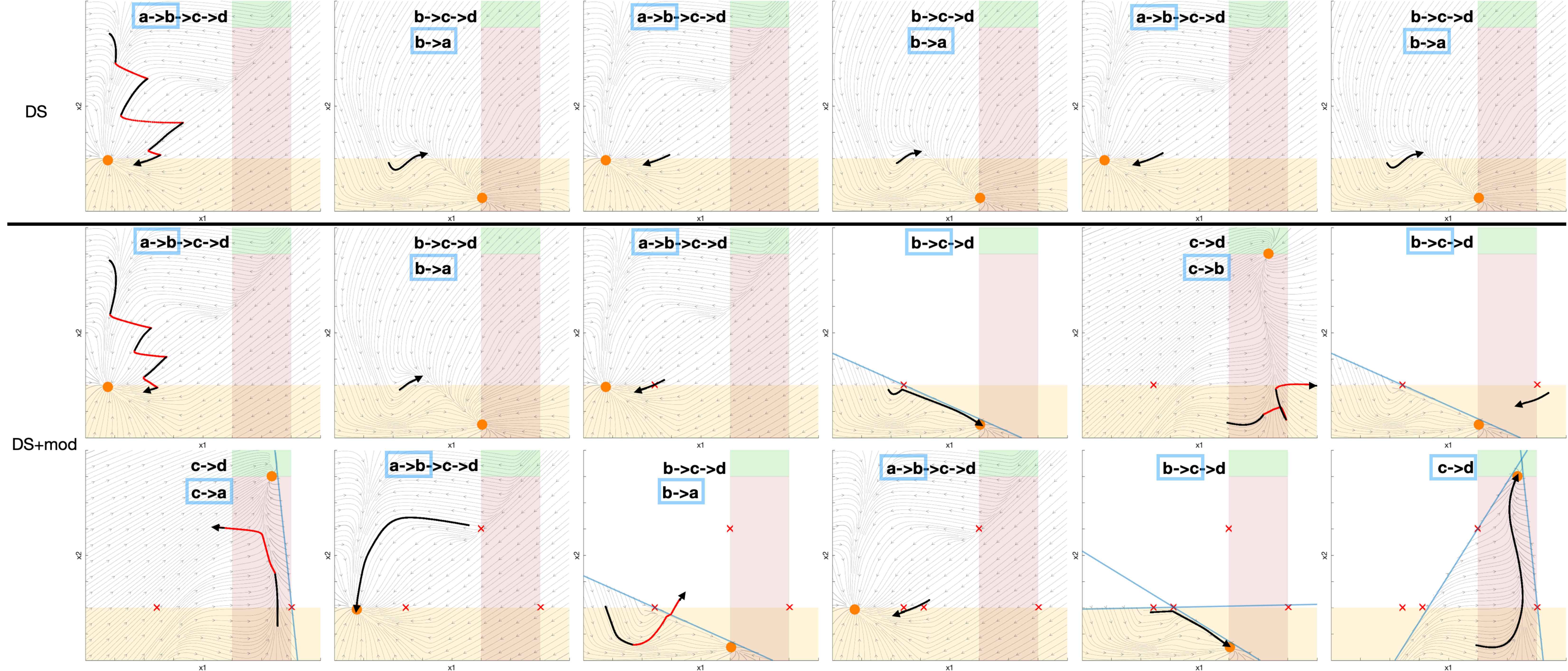}
\caption{\footnotesize Rollouts of a multi-step scooping task under perturbations. The first row shows that sequencing DS policies with an automaton can lead to looping without boundary estimation. The second and third rows show that modulation prevents looping and enables the system to eventually reach the goal mode despite repeated perturbations. We show the desired discrete plan at the top of each sub-figure and annotate the current mode transition detected in the blue box. Black and red trajectories signify original and perturbed rollouts. \label{fig:modulation} \vspace{-15pt}}
\end{figure}

\subsection{Multi-Modal Reactivity and Generalization to New Tasks}\label{sec:scooping_soup_task}
\vspace{-8.5pt}
We now empirically demonstrate that having a reactive discrete plan alone is insufficient to guarantee task success without mode invariance for tasks with multiple modes. Consider the multi-modal soup-scooping task introduced in Fig. \ref{fig:intro}. Formally, we define three environment APs, $r, s, t$, sensing the spoon is in contact with the soup, has soup on it, and has arrived at a target location respectively. Given successful demonstrations, sensors will record discrete transitions $(\lnot r\land \lnot s \land \lnot t) \Rightarrow (r\land \lnot s \land \lnot t) \Rightarrow (\lnot r \land s \land \lnot t) \Rightarrow (\lnot r \land \lnot s \land t)$, from which four unique sensor states are identified. We label each sensor state as a mode with robot AP $a \text{ (reaching) }\Rightarrow b \text{ (scooping) } \Rightarrow c \text{ (transporting) } \Rightarrow d \text{ (done)}$. The Invariance of mode $b$ enforces contact with soup during scooping, and the invariance of mode $c$ constrains the spoon's orientation to avoid spilling. We follow the TLP convention to assume LTL formulas are provided by domain experts (although they can also be learned from demonstrations \cite{shah2018bayesian, chou2021learning}.) The specific LTL for the soup-scooping task is detailed in Appendix \ref{sec:multi-modal}, and can be converted into a task automaton as shown in Fig. \ref{fig:method}. One might assume the automaton is sufficient to guarantee task success without modulation, as it only needs to replan a finite number of times assuming a finite number of perturbations; however, not enforcing mode invariance can lead to looping at the discrete level, and ultimately renders the goal unreachable, as depicted in the top row of Fig. \ref{fig:modulation}. In contrast, looping is prevented when modulation is enabled, as the system experiences each invariance failure only once. 


\textbf{Robot Experiments} First, we implement the soup-scooping task on a Franka Emika robot arm as shown in Fig.~\ref{fig:robot_intro}. We show in videos on our website that (1) DS allows our system to compliantly react to motion-level perturbations while ensuring system stability; (2) LTL allows our system to replan in order to recover from task-level perturbations; and (3) our modulation ensures the robot learns from previous invariance failures to avoid repeating them. To test robustness against unbiased perturbations, we collect 30 trials from 6 humans as seen in Appendix \ref{sec:robot_scoop}. All trials succeed eventually in videos. We do not cherry-pick these results, and the empirical 100\% success rate further corroborates our theoretic success guarantee. Second, we implement an inspection task as a permanent interactive exhibition at \href{https://yanweiw.github.io/tli/#Museum}{MIT Museum}, with details documented in Appendix \ref{sec:inspection}. Lastly, we show a color tracing task testing different automaton structures with details in Appendix \ref{sec:color}.

\textbf{Generalization} LTL-DS can generalize to a new task by reusing learned DS if the new LTL shares the same set of modes. Consider another multi-step task of adding chicken and broccoli to a pot. Different humans might give demonstrations with different modal structures (e.g., adding chicken first vs adding broccoli first). LTL-DS can be reformulated to learn a policy for each mode transition (each mode can now have multiple policies), resulting in a collection of DS skills that can be flexibly recombined to solve new tasks. To generate different task LTLs, a human only needs to edit the $\phi_t^s$ portion of the original LTL formula. We provide further details of this analysis in Appendix \ref{sec:generalization}.
\vspace{-12.5pt}

\section{Limitations} 
\vspace{-8.5pt}
TLI assumes the existence of suitable mode abstractions, reactive logic formulas and perfect sensors to detect mode transitions, which can be difficult to obtain without non-trivial domain knowledge. Our work is based on the assumption that for well-defined tasks (e.g., assembly tasks in factory settings), domain expertise in the form of a logic formula is a cheaper knowledge source than collecting hundreds of motion trajectories to avoid covariate shift (we use up to $3$ demonstrations in all experiments). Moreover, even when abstractions for a task are given by an oracle, an LfD method without either the invariance or the reachability property will not have a formal guarantee of successful task replay, which is this work's focus. In future work, we will learn mode abstractions directly from sensor streams such as videos so that our approach gains more autonomy without losing reactivity. 


\vspace{-12.5pt}

\section{Conclusion}
\label{sec:conclusion}
\vspace{-8.5pt}
In this paper, we formally introduce the problem of \textit{temporal logic imitation} as imitating continuous motions that satisfy an LTL specification. We identify the fact that learned policies do not necessarily satisfy the bisimulation criteria as the main challenge of applying LfD methods to multi-step tasks. To address this issue, we propose a DS-based approach that can iteratively estimate mode boundaries to ensure invariance and reachability. Combining the task-level reactivity of LTL and the motion-level reactivity of DS, we arrive at an imitation learning system able to robustly perform various multi-step tasks under arbitrary perturbations given only a small number of demonstrations. We demonstrate our system's practicality on a real Franka robot. 


\section*{Acknowledgments}

We would like to thank Jon DeCastro, Chuchu Fan, Terry Suh, Rachel Holladay, Rohan Chitnis, Tom Silver, Yilun Zhou, Naomi Schurr, and Yuanzhen Pan for their invaluable advice and help.

\bibliography{reference}  
\clearpage

\appendix
\section{Proofs}\label{sec:proof}
\setcounter{theorem}{0}
\begin{theorem}
    (Key Contribution 1) A nonlinear DS defined by Eq. \ref{eq:ds_eq}, learned from demonstrations, and modulated by cutting planes as described in Section \ref{sec:invariance} with the reference point $x^r$ set at the attractor $x^*$, will never penetrate the cuts and is G.A.S. at $x^*$.
\end{theorem} 
\textit{Proof\quad} Let the region bounded by cuts be $\mathcal{D}$, which is non-empty as it contains at least one demonstration. If $x \notin \mathcal{D}$, i.e., $x$ is outside the cuts, the nominal DS $f(x)$ will not be modulated. Since $f(x)$ is G.A.S. at $x^*$ and $x^* \in \mathcal{D}$, a robot state at $x$ will enter $\mathcal{D}$ in a finite amount of time. If $x \in \mathcal{D}$ and $[E(x)^{-1}f(x)]_1<0$, which corresponds to $f(x)$ having a negative component in the direction of $\bold{r^*}(x)=\frac{x-x^*}{\|x-x^*\|}$, $f(x)$ is moving away from cuts and toward the attractor. In this case, we leave $f(x)$ unmodulated and the original G.A.S. property holds true. If $x \in \mathcal{D}$ and $[E(x)^{-1}f(x)]_1\geq0$, where the nominal DS could flow out of the cuts, we apply modulation---and, by construction, $M(x)f(x)$ stays inside the cuts. To prove the stability of the modulated DS, we show that the Lyapunov candidate, $V(x)=(x-x^*)^T P(x-x^*)$, in satisfying $\dot{V}(x)=\frac{\partial V(x)}{\partial x}f(x) < 0$ for $f(x)$, also satisfies the Lyapunov condition for $M(x)f(x)$ (omitting matrix dependency upon $x$ to reduce clutter):
\begin{equation}
\begin{aligned}
\dot{V}(x) &= \frac{\partial V(x)}{\partial x}Mf(x) = \frac{\partial V(x)}{\partial x}EDE^{-1}f(x)\\
& = \frac{\partial V(x)}{\partial x}E~\textbf{diag(}1-\Gamma(x),1,...,1\textbf{)}E^{-1}f(x)\\
& = \frac{\partial V(x)}{\partial x}f(x) - \frac{\partial V(x)}{\partial x}E~\textbf{diag}(\Gamma(x),0,...,0)E^{-1}f(x)\\
& < 0 - \frac{\partial V(x)}{\partial x}\bold{r^*}(x)^T\Gamma(x)[E^{-1}f(x)]_1\\
& < 0 - \underbrace{2(x-x^*)^T P \frac{x-x^*}{\|x-x^*\|}}_{>0 ~\text{as}~ P\succ 0}\underbrace{\Gamma(x)}_{>0}\underbrace{[E^{-1}f(x)]_1}_{\geq0} < 0
\end{aligned}
\end{equation}
Therefore, $M(x)f(x)$ is G.A.S. \qedsymbol{}\\

The following lemmas support the proof of \textbf{Theorem 2}.
\begin{lemma} \label{eq:lemma1}
    LTL-DS generates a discrete reactive plan of mode sequence that satisfies any LTL formula provided to the algorithm. 
\end{lemma}
\textit{Proof\quad} Since the task automaton is converted from an LTL formula, all resulting discrete plans of mode sequence (including the replanned sequence caused by perturbations) are correct by construction as long as the environment is admissible. \qedsymbol{}

\begin{lemma} \label{eq:lemma2}
    If a mode transition $\sigma_i \Rightarrow \sigma_j$ has been observed in the demonstrations, $\sigma_j$ is reachable from $\sigma_i$ by DS $f_{i}$.
\end{lemma}
\textit{Proof\quad} Since $\sigma_i \Rightarrow \sigma_j$ has been demonstrated, $\sigma_i$ and $\sigma_j$ must be connected; let them share a guard, $G_{ij}$. Assigning a globally stable DS $f(\cdot):\mathbb{R}^n\rightarrow \mathbb{R}^n$ to each mode $\sigma_i$ with region $\delta_i \subset \mathbb{R}^n$ guarantees asymptotic convergence of all $x$ in $\delta_i$ to the attractor, $x^*_{i}$ by DS $f_{i}$. Placing $x^*_{i}$ on guard $G_{ij}$ ensures that $x^*_{i} \in \delta_j$, and thus $\forall s ~\sigma_i \Rightarrow \sigma_j ~\text{as} ~x \rightarrow x^*_{i}$. \qedsymbol{}

\begin{lemma} \label{eq:lemma3}
    If an unseen mode transition $\sigma_i \Rightarrow \sigma_j$ occurs unexpectedly, the system will not be stuck in $\sigma_j$.
\end{lemma}
\textit{Proof\quad} While the transition $\sigma_i \Rightarrow \sigma_j$ has not been seen in demonstrations, Asm. \ref{eq:non-oracle} ensures that mode $\sigma_j$ has been observed and its associated DS $f_{j}$ has been learned. Since the LTL GR(1) fragment does not permit clauses in the form of (\textbf{F G$\phi$}), which states $\phi$ is eventually globally true (i.e., the system can stay in $\sigma_j$ forever), every discrete plan will have to in finite steps result in $\sigma_j \Rightarrow \sigma_k$ for some $k,\ j\neq k$. Having learned $f_{j}$ also validates the existence of $x^*_{j}$---and, thus, a continuous trajectory toward $G_{jk}$. \qedsymbol{}

\begin{theorem} (Key Contribution 2) \label{eq:kc2}
    The continuous trace of system states generated by LTL-DS satisfies any LTL specification $\phi$ under Asm. \ref{eq:convexity}, \ref{eq:finiteness}, and \ref{eq:non-oracle}. 
\end{theorem}

\textit{Proof\quad} Lemma \ref{eq:lemma1} proves any discrete plan generated by LTL-DS satisfies the LTL specification. Lemmas \ref{eq:lemma2} and \ref{eq:lemma3} and Asm. \ref{eq:finiteness} ensure the reachability condition of all modes. Thm. \ref{eq:kc1} certifies the modulated DS will be bounded inside the cuts, and thus the mode these cuts inner-approximate. Consequently, a finite number of external perturbations only require a finite number of cuts in order to ensure mode invariance. Given that bisimulation is fulfilled, the continuous trace generated by LTL-DS simulates a LTL-satisficing discrete plan, and thus satisfies the LTL. \qedsymbol{}

\section{Motivation for Mode-based Imitation}
Our work aims to achieve generalization in regions of the state space not covered by initial demonstrations. A well-studied line of research is to collect more expert data \cite{ross2011reduction} so that the policy will learn to recover from out-of-distribution states. Our central observation in Fig. \ref{fig:intro} is that there exists some threshold that separates trajectories deviating from expert demonstrations (black) into successes (blue) and failures (red). The threshold can be embodied in mode boundaries, which lead to the notion of a discrete mode sequence that acts as the fundamental success criteria for any continuous motions. In fact, online data collection to improve policies in DAGGER \cite{ross2011reduction} can be seen as implicitly enforcing mode invariance. We take the alternative approach of explicitly estimating mode boundaries and shift the burden from querying for more data to querying for a task automaton in the language of LTL.

\section{Sensor-based Motion Segmentation and Attractor Identification} \label{sec:segmentation}

\begin{table}[!h]
    \centering
    \begin{tabular}{ccccccccccc}
    \hline
    time step & 1 & 2 & 3 & 4 & 5 & 6 & 7 & 8 & 9 & 10 \\ \hline
    demo 1    & \color{red}$x^{1,1}$ & \color{red}$x^{2,1}$ & \color{blue}$x^{3,1}$ & \color{blue}$x^{4,1}$ & \color{blue}$x^{5,1}$ & \color{blue}$x^{6,1}$ & \color{green}$x^{7,1}$ & \color{green}$x^{8,1}$ & \color{green}$x^{9,1}$ & \color{green}$x^{10,1}$\\
    demo 2    & \color{red}$x^{1,2}$ & \color{red}$x^{2,2}$ & \color{red}$x^{3,2}$ & \color{red}$x^{4,2}$ & \color{blue}$x^{5,2}$ & \color{blue}$x^{6,2}$ & \color{blue}$x^{7,2}$ & \color{blue}$x^{8,2}$ & \color{blue}$x^{9,2}$ & \color{green}$x^{10,2}$\\
    \hline
    \end{tabular}
    \caption{Demonstrations are segmented into three AP regions (shown by color) based on three unique sensor states for DS learning. We use the average location of the last states (transition states to the next AP) in each AP as the attractor for the corresponding DS.}
    \label{tab:segmentation}
\end{table}

Let $\{\{x^{t,k},\dot{x}^{t,k}, \alpha^{t,k}\}_{t=1}^{T_k}\}_{k=1}^{K}$ be $K$ demonstrations of length $T_k$. The motion trajectories in $x^{t, k}$ are clustered and segmented into the same AP region if they share the same sensor state $\alpha^{t,k}$. For example, in Table. \ref{tab:segmentation} two demonstrations of ten time steps form three AP regions (colored by red, blue and green) based on three unique sensor readings. To obtain the attractor for each of the three DS to be learned , we average the last state in the trajectory segment. For instance, the average location of $x^{2,1}$ and $x^{4,2}$, $x^{6,1}$ and $x^{9,2}$, $x^{10,1}$ and $x^{10,2}$ become the attractor for the red, blue and green AP's DS respectively. 

\section{Relation of TLI to Prior Work}
\begin{figure}[!h]
  \centering
  \includegraphics[width=0.8\linewidth]{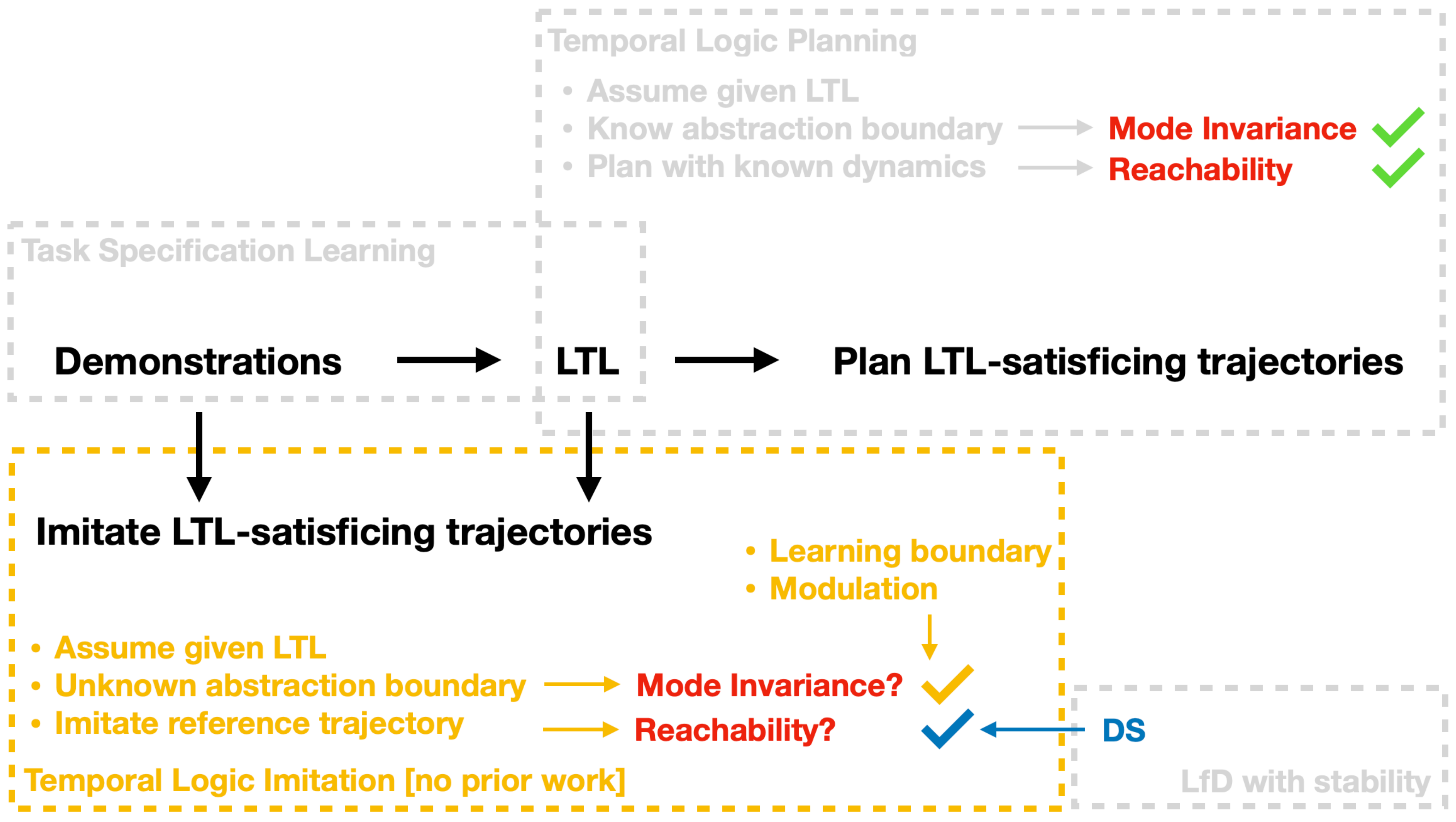}
  \caption{How TLI fits in relation to prior work, where gray dashed boxes represent prior work and yellow dashed box highlights our contribution.}
  \label{fig:prior_work}
\end{figure}

This work explores a novel LfD problem formulation (temporal logic imitation) that is closely related to three research communities. First, there is a large body of work on learning task specifications in the form of LTL formulas from demonstrations \cite{shah2018bayesian, chou2021learning, kasenberg2017interpretable}. We do not repeat their endeavor in this work and assume the LTL formulas are given. Second, given LTL formulas there is another community (temporal logic planning) that studies how to plan a continuous trajectory that satisfies the given LTL \cite{belta2007symbolic, wolff2013automaton, plaku2016motion, kress2018synthesis}. Their assumption of known abstraction boundaries and known dynamics allow the planned trajectory to satisfy the invariance and reachability (bisimulation) criteria respectively, thus certifying the planned continuous trajectory will satisfy any discrete plan. Our observation is that the bisimulation criteria can also be used to certify that a LfD policy can simulate the discrete plan encoded by any LTL formula, which we dub as the problem of TLI. To the best of our knowledge, our work is the first to formalize TLI and investigate its unique challenges inherited from the LfD setting. On the one hand, we no longer have dynamics to plan with but we have reference trajectories to imitate. To satisfy reachability, it is necessary to leverage a third body of work--LfD methods with global stability guarantee (DS) \cite{khansari2011learning, figueroa2018physically, billard2022learning}. On the other hand, we note LfD methods typically do not satisfy mode invariance due to unknown mode boundaries that are also innate to the LfD setting. Thus, we propose learning an approximate mode boundary leveraging sparse sensor events and then modulating the learned policies to be mode invariant. We prove DS policies in particular after modulation will still satisfy reachability, and consequently certifying they will satisfy any given LTL formulas. Figure \ref{fig:prior_work} summarizes how TLI's relationship to prior work, where gray dashed boxes represent prior work and yellow dashed box highlights our contribution.

\section{Single-mode Experiments} \label{sec:single-mode}
\subsection{Experiment Details}
\begin{figure*}[h!] 
    \centering
  \begin{minipage}[b]{\textwidth}
    \includegraphics[width=\textwidth]{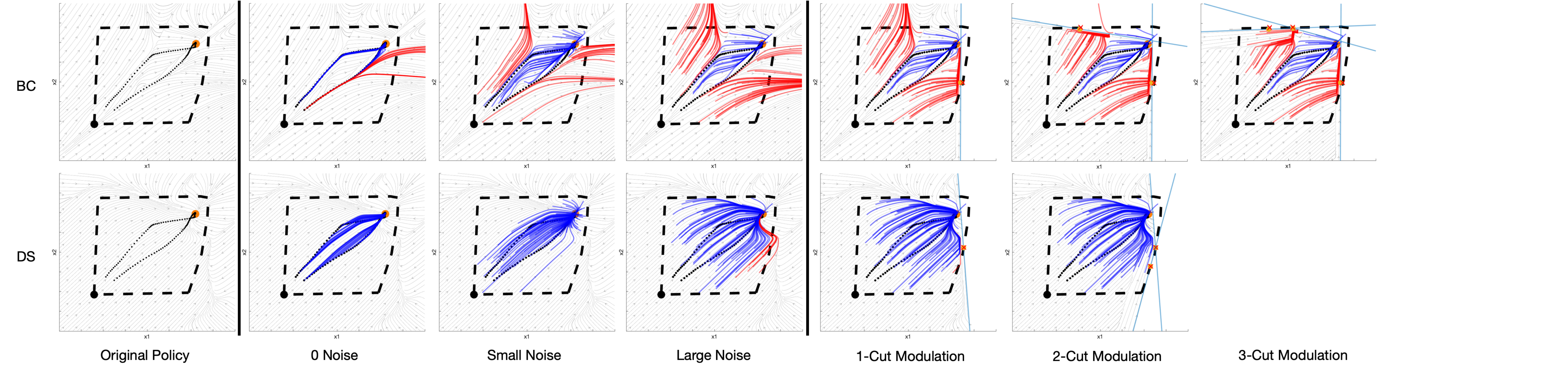}
  \end{minipage}
  \hfill
  \begin{minipage}[b]{\textwidth}
    \includegraphics[width=\textwidth]{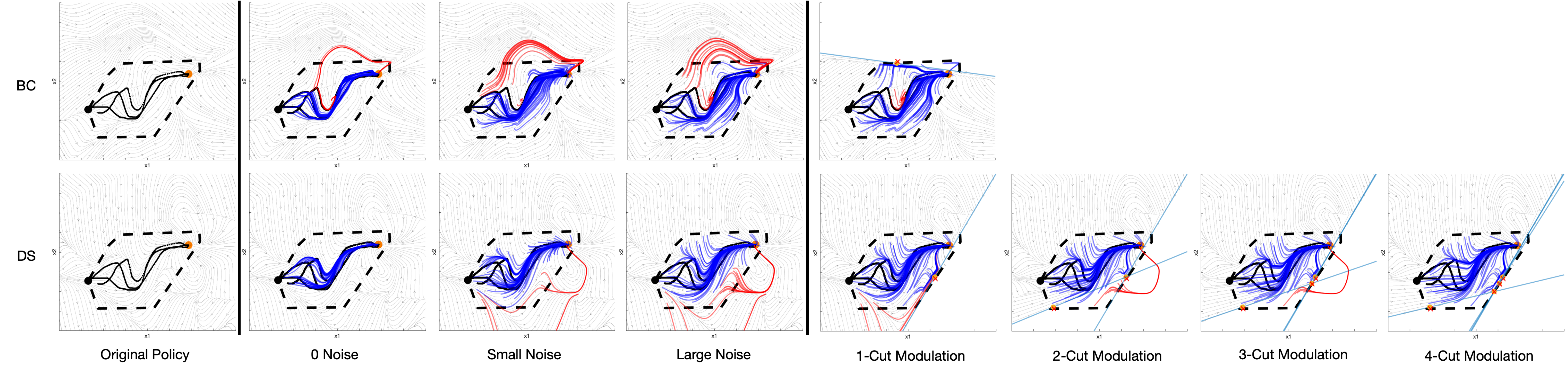}
  \end{minipage}
    \hfill
  \begin{minipage}[b]{\textwidth}
    \includegraphics[width=\textwidth]{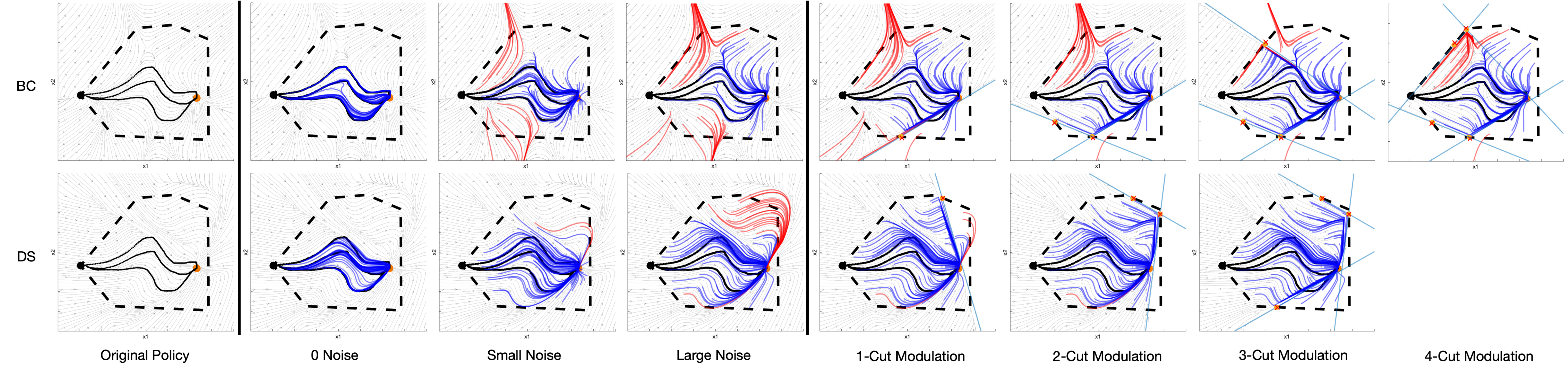}
  \end{minipage}  
  \hfill
  \begin{minipage}[b]{\textwidth}
    \includegraphics[width=\textwidth]{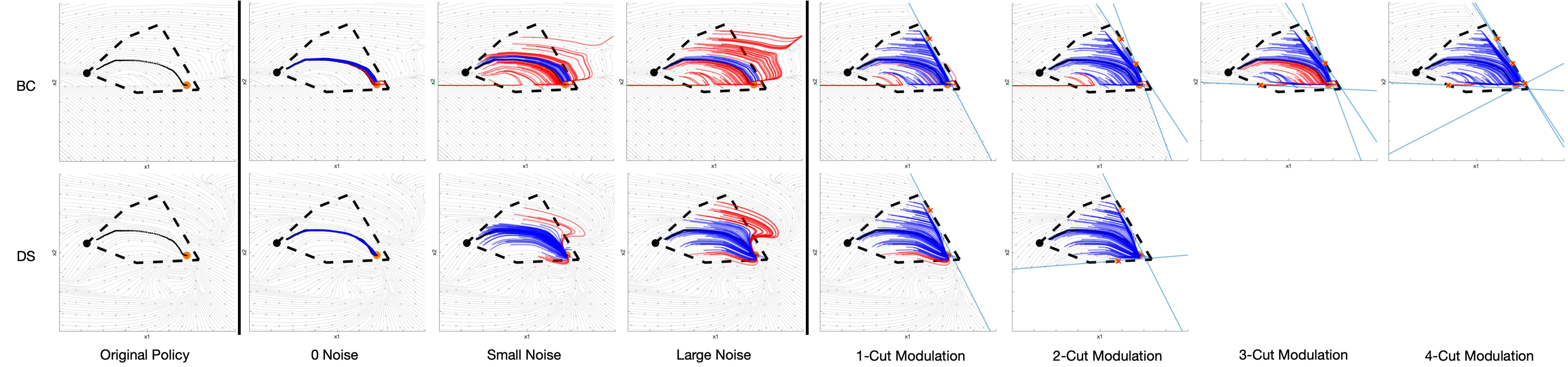}
  \end{minipage}
  \begin{minipage}[b]{\textwidth}
    \includegraphics[width=\textwidth]{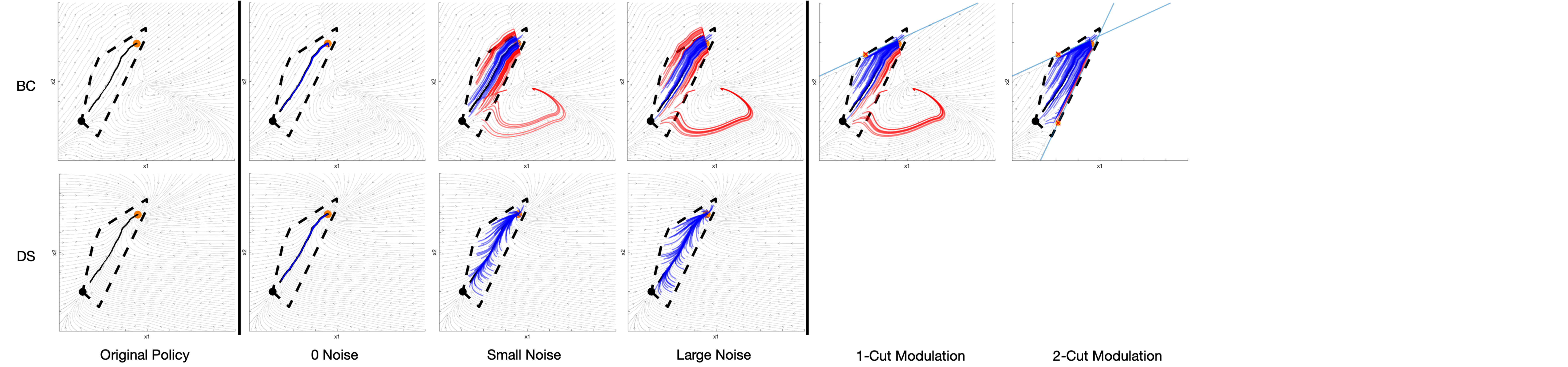}
  \end{minipage}
  \begin{minipage}[b]{\textwidth}
    \caption{\footnotesize For each convex mode, we use 1-3 demonstrations for learning shown in black. Successful rollouts are shown in blue while unsuccessful rollouts are shown in red. We apply modulation to the large noise case and within four cuts all DS policies are modulated to be mode invariant. While BC policies can also be modulated to be mode invariant, they still suffer from existing reachability failures prior to modulation as well as new reachability failures introduced by modulation. For example, in BC flows that are originally flowing out of the mode can lead to spurious attractors at the cuts after modulation. We prove this will not happen for DS due to its stability guarantee.} \label{fig:additional_rollouts}
  \end{minipage} 
\end{figure*}

We randomly generate convex modes and draw $1-3$ human demonstrations, as seen in Fig. \ref{fig:rollout} (left). To check mode invariance, we sample starting states from the demonstration distribution perturbed by Gaussian noise with standard deviation of 0\%, 5\%, and 30\% of the workspace dimension. Sampling with zero noise corresponds to sampling directly on the demonstration states, and sampling with a large amount of noise corresponds to sampling from the entire mode region. To enforce invariance, we iteratively sample a failure state and add a cut until all invariance failures are corrected. A task replay is successful if and only if an execution trajectory both reaches the goal and stays within the mode. For each randomly generated convex mode, we sampled 100 starting states and computed the average success rate for 50 trials. We show DS+modulation ensures both reachability and invariance for five additional randomly sampled convex modes in Fig. \ref{fig:additional_rollouts}.
\subsection{BC Policy Architecture and Training Details}
For the Neural-network-based BC policy, we use an MLP architecture that consists of 2 hidden layers both with 100 neurons followed by ReLU activations. We use tanh as the output activation, and we re-scale the output from tanh layer to [-50 -- 50]. Each demonstration trajectory consists of about 200 pairs of states and velocities as the training data to the network. Since we are training a state-based policy that predicts velocities from input states, we treat these data points as i.i.d. For training, we use Adam as the optimizer with a learning rate of 1e-3 for max 5000 epochs. 
\subsection{QCQP Optimization Details} \label{sec:qcqp}
To find the normal direction of a hyperplane that goes through a last-in-mode states $x^{T_f-1}$ in Sec. \ref{sec:invariance}, we solve the following optimization problem, where $w$ is the normal direction we are searching over; $f=1, 2, ...$ indexes a set of failure states; and $T_f$ is the corresponding time-step of first detecting an invariance failure ($T$ alone is used as matrix transpose.)
\begin{equation}
\begin{aligned}
\min_{w} \quad & (w^T(x^*-x^{T_f-1}))^2\\
\textrm{s.t.} \quad & \|w\|=1 \\
& w^T(x^*-x^{T_f-1}) \leq 0\\
& w^T(x^0-x^{T_f-1}) \leq 0\\
& w^T(x^{T_{f^{\prime}}-1} - x^{T_f-1}) \leq 0 \quad \forall f^{\prime}
\end{aligned}
\end{equation}
While specialized QCQP packages can be used to solve this optimization problem, we use a generic nonlinear Matlab function \texttt{fmincon} to solve for $w$ in our implementation.

\section{Multi-modal Experiments} \label{sec:multi-modal}
After abstractions---environment APs, $r,s,t$, and robot APs, $a,b,c,d$---for the soup-scooping task in Sec. \ref{sec:scooping_soup_task} are defined, the reactive LTL formula can be written as $\phi = ((\phi^e_i\land \phi^e_t\land \phi^e_g) \rightarrow (\phi^s_i\land \phi^s_t\land \phi^s_g))$. $\phi_i^s$ and $\phi_i^e$ specify the system's initial mode, $a$, and the corresponding sensor state. $\phi_g^s$ and $\phi_g^e$ set mode $d$ as the eventual goal for the robot, with no particular goal for the environment. $\phi_t^e$ specifies the environmental constraints that determine which sensor states are true in each mode, as well as the fact that the system can only be in one mode at any times. $\phi_t^s$ specifies all valid transitions for each mode.
\begin{align*}
    \phi_i^e = &\ \lnot r\land \lnot s \land \lnot t; \quad \phi_g^e = True;\\
    \phi_t^e = &\ \textbf{G}((a \leftrightarrow (\lnot r\land \lnot s \land \lnot t)
                \land (b \leftrightarrow (r\land \lnot s \land \lnot t)) \\
                &\quad \land (c \leftrightarrow (\lnot r \land s \land \lnot t)
                \land (d \leftrightarrow (\lnot r \land \lnot s \land t)) \\
                & \land \textbf{G}((a\land \lnot b \land \lnot c \land \lnot d) \lor               (\lnot a\land b \land \lnot c \land \lnot d)\\
               & \quad \lor (\lnot a\land \lnot b \land c \land \lnot d)
               \lor (\lnot a\land \lnot b \land \lnot c \land d));\\
    \phi_i^s = &\ a; \quad \phi_g^s = \textbf{GF}d;\\
    \phi_t^s = &\ \textbf{G}((a \rightarrow (\textbf{X}a \lor \textbf{X}b))
               \land (b \rightarrow (\textbf{X}a \lor \textbf{X}b \lor \textbf{X}c))\\
               & \quad \land (c \rightarrow (\textbf{X}a \lor \textbf{X}b \lor \textbf{X}c \lor \textbf{X}d))
               \land (d \rightarrow \textbf{X}d)); 
\end{align*}
 \textbf{Automatic construction of GR(1) LTL formulas} One benefit of using the GR(1) fragment of LTL is that it provides a well-defined template for defining a system's reactivity \cite{kress2009temporal}\footnote{The main benefit is that GR(1) formulas allow fast synthesis of their equivalent automatons \cite{piterman2006synthesis}.}. While in this work we follow the TLP convention that assumes the full GR(1) formulas are given, the majority of these formulas can actually be automatically generated if Asm. \ref{eq:non-oracle} holds true. Specifically, once the abstraction, $r, s, t, a, b, c, d$ is defined, formulas $\phi_t^e, \phi_g^e$ are correspondingly defined as shown above, and they remain the same for different demonstrations. If a demonstration displaying $a\Rightarrow b\Rightarrow c\Rightarrow d$ is subsequently recorded, formulas $\phi_i^e, \phi_i^s, \phi_g^s$ as shown above can then be inferred. Additionally a partial formula $\phi_t^s=\textbf{G}((a \rightarrow (\textbf{X}a \lor \textbf{X}b)) \land (b \rightarrow (\textbf{X}b \lor \textbf{X}c) \land (c \rightarrow (\textbf{X}c \lor \textbf{X}d)) \land (d \rightarrow \textbf{X}d))$ can be inferred. Synthesis from this partial $\phi = ((\phi^e_i\land \phi^e_t\land \phi^e_g) \rightarrow (\phi^s_i\land \phi^s_t\land \phi^s_g))$ results in a partial automaton in Fig. \ref{fig:method} with only black edges. During online iteration, if perturbations cause unexpected transitions, $b\Rightarrow a$ and/or $c\Rightarrow a$ and/or $c\Rightarrow b$, which are previously not observed in the demonstration, $\phi_t^s$ will be modified to incorporate those newly observed transitions as valid  mode switches, and a new automaton will be re-synthesized from the updated formula $\phi$. The gray edges in Fig. \ref{fig:method} reflect those updates after invariance failures are experienced. Asm. \ref{eq:non-oracle} ensures the completeness of the demonstrations with respect to modes, i.e., the initially synthesized automaton might be missing edges but not nodes compared to an automaton synthesized from the ground-truth full formula. For general ground-truth LTL formulas not part of the GR(1) fragment or demonstrations not necessarily satisfying Asm. \ref{eq:non-oracle}, we cannot construct the formulas using the procedure outlined above. In that case, we can learn the formulas from demonstrations in a separate stage \cite{shah2018bayesian, chou2021learning}. 
 
 In this work, we assume full LTL formulas are provided by domain experts. Since they are full specifications of tasks, the resulting automatons will be complete w.r.t. all valid mode transitions (e.g., including both the black and gray edges in Fig. \ref{fig:method}), and will only need to be synthesized once. Given the soup-scooping LTL defined above, we ran $10$ experiments, generating $1-3$ demonstrations for each, and learning a DS per mode. We applied perturbations uniformly sampled in all directions of any magnitudes up to the dimension of the entire workspace in order to empirically verify the task-success guarantee. We follow the QCQP optimization defined in Appendix B to find cuts to modulate the DS. Simulation videos can be found on the project page.

\section{Generalization Results} \label{sec:generalization}
\begin{figure}[h] 
  \centering
  \includegraphics[width=1\linewidth]{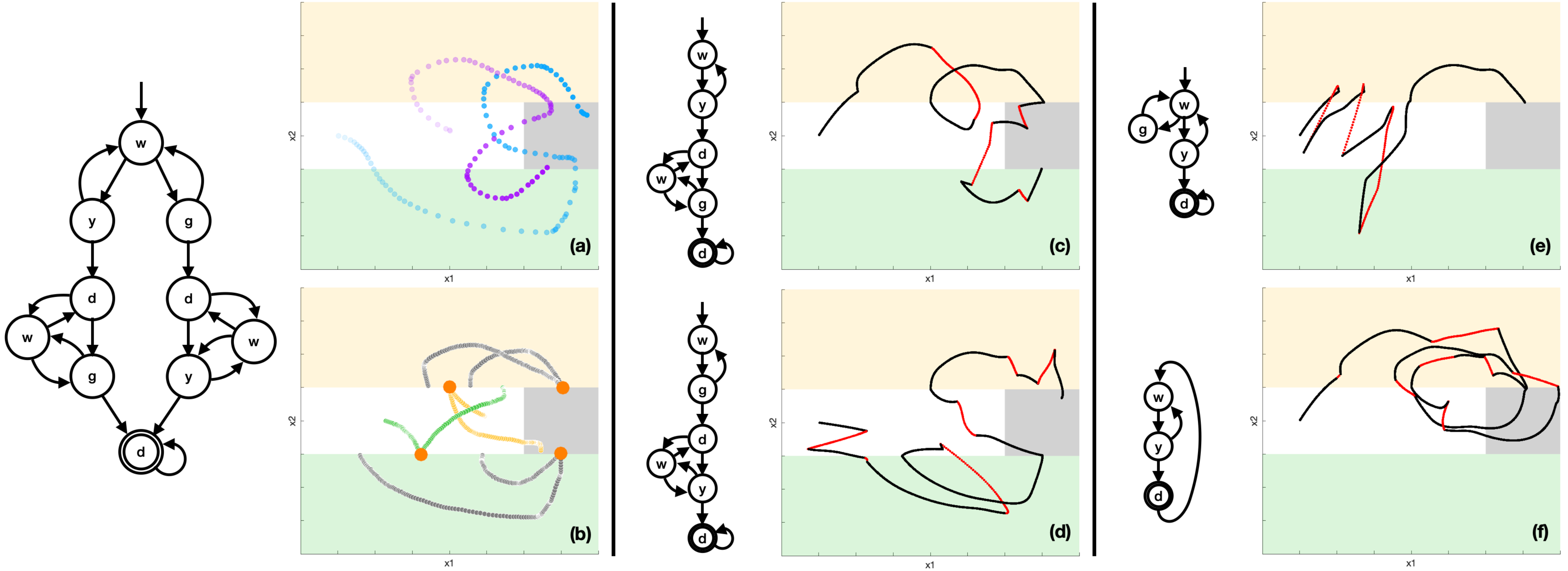}
  \caption{\footnotesize Reuse learned DS for new tasks by editing their task automaton via LTL. \textbf{(a)} Two demonstrations (purple and blue) solve a task of adding both chicken (yellow region) and broccoli (green region) to a pot (dark region) in different order as visualized by the automaton on the left. \textbf{(b)} Demonstrations are segmented into parts by mode region, which are then shifted to their average attractor locations (orange circles) for DS learning. Since a mode can proceed to different next modes in this case, we adapt the mode-based DS policy to be associated with each unique mode transition instead. The learned DS can generalize to new tasks given new automatons specifying \textbf{(c)} the order: white (w)  $\Rightarrow$ yellow (y) $\Rightarrow$ dark (d) $\Rightarrow$ green (g) $\Rightarrow$ dark (d), \textbf{(d)} the order: white $\Rightarrow$ green $\Rightarrow$ dark $\Rightarrow$ yellow $\Rightarrow$ dark, \textbf{(e)} the goal: getting only chicken, or \textbf{(f)} the goal: getting chicken continuously. The trajectory in red shows perturbations and the trajectory in black shows recovery according to the automaton structure. 
  }
  \label{fig:generalization}
\end{figure}

LTL-DS can generalize to a new task by reusing learned DS if the new LTL shares the same set of modes. Consider another multi-step task of adding chicken and broccoli to a pot. Different humans might give demonstrations with different modal structures (e.g., adding chicken vs adding broccoli first) as seen in Fig. \ref{fig:generalization} \textbf{(a)}. LTL-DS learns individual DS which can be flexibly combined to solve new tasks with new task automatons, as illustrated in Fig. \ref{fig:generalization} \textbf{(c-f)}. To get these different task automatons, a human just needs to edit the $\phi_t^s$ portion of the LTL formulas differently.

We describe LTL formulas for variants of the cooking task of adding chicken and broccoli to a pot as visualized in Fig. \ref{fig:generalization}. We use mode AP $w$, $y$, $g$, $d$ to define configurations of empty spoon (white region), transferring chicken (yellow region), transferring broccoli (green region), and finally dropping food in the pot (dark region) . We follow the description of scooping task LTL to define $\phi^e_i, \phi^e_t, \phi^e_g, \phi^s_i, \phi^s_g$ for the cooking tasks, which are shared by them all. We focus on $\phi^s_t$ here as it captures mode transitions and is different for a different task. We denote the $\phi^s_t$ portion of LTL for the new task of adding chicken first, adding broccoli first, adding chicken only, and adding chicken continuously as $\phi_{cb}$, $\phi_{bc}$, $\phi_{c}$, and $\phi_{cc}$ respectively.
\begin{align*}
        \phi_{cb} = &\ \textbf{G}((w_1 \rightarrow \textbf{X}y) \land (y \rightarrow (\textbf{X}w_1 \lor \textbf{X}d_1)) \\ 
        & \land (d_1 \rightarrow (\textbf{X}w_2 \lor \textbf{X}g)) \land (w_2 \rightarrow \textbf{X}g) \\
        & \land (g \rightarrow (\textbf{X}w_2 \lor \textbf{X}d_2)) \land (d_2 \rightarrow \textbf{X}d_2)); \\
        \phi_{bc} = &\ \textbf{G}((w_1 \rightarrow \textbf{X}g) \land (g \rightarrow (\textbf{X}w_1 \lor \textbf{X}d_1)) \\ 
        & \land (d_1 \rightarrow (\textbf{X}w_2 \lor \textbf{X}y)) \land (w_2 \rightarrow \textbf{X}y) \\
        & \land (y \rightarrow (\textbf{X}w_2 \lor \textbf{X}d_2)) \land (d_2 \rightarrow \textbf{X}d_2)); \\
        \phi_{c} = &\ \textbf{G}((w \rightarrow (\textbf{X}y \lor \textbf{X}g)) \land (y \rightarrow (\textbf{X}w \lor \textbf{X}d)) \\ 
        & \land (g \rightarrow \textbf{X}w) \land (d \rightarrow \textbf{X}d)); \\
        \phi_{cc} = &\ \textbf{G}((w \rightarrow \textbf{X}y) \land (y \rightarrow (\textbf{X}w \lor \textbf{X}d)) \land (d \rightarrow \textbf{X}w)); 
\end{align*}
Note mode $w_1$ and $w_2$ denote visiting the white region before and after some food has been added to the pot and they share the same motion policy. The same goes for mode $d_1$ and $d_2$. These formulas can be converted to task automatons in Fig. \ref{fig:generalization}. We show animations of these tasks on the project page.  

\section{Robot Experiment 1: Soup-Scooping} \label{sec:robot_scoop}
\begin{figure}[!h]
  \centering
  \includegraphics[width=0.6\linewidth]{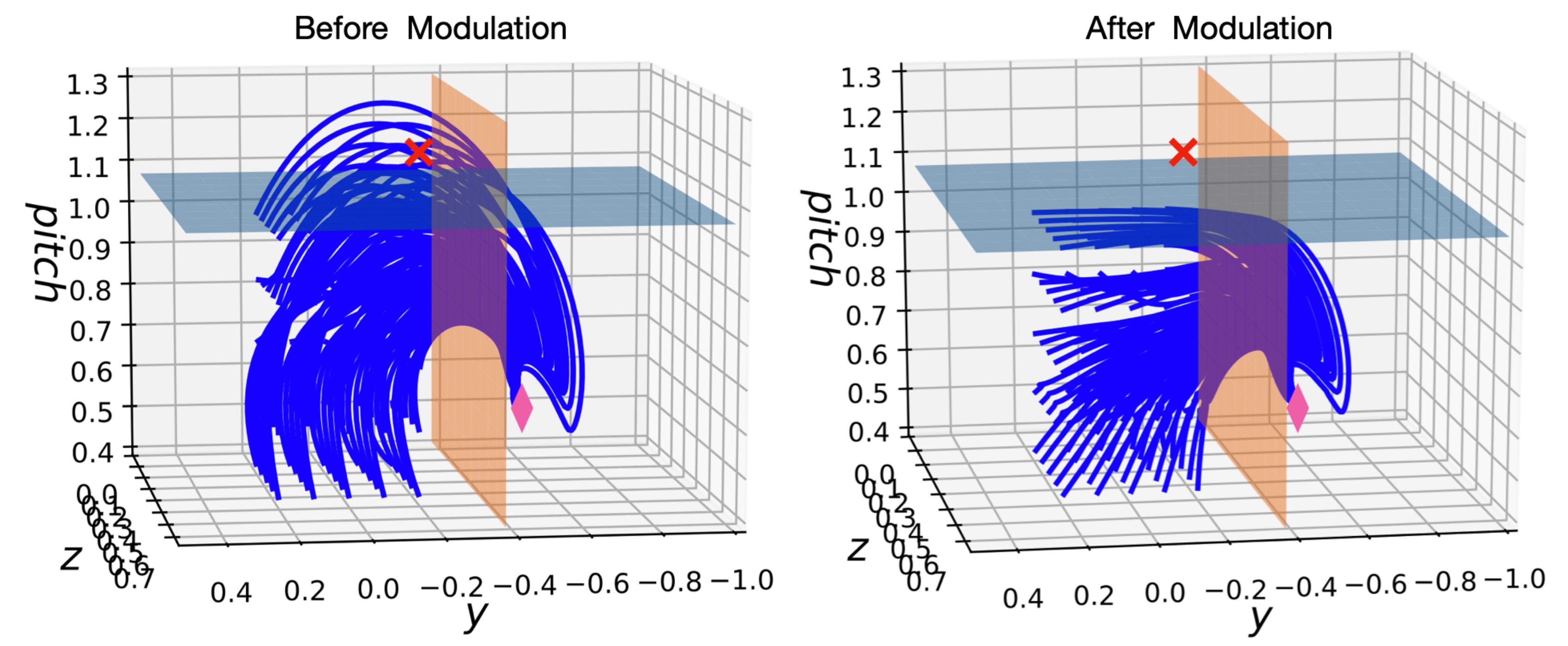}
  \caption{\footnotesize Modulation of sampled robot trajectories. The orange plane represents a guard surface between the transporting mode and the done mode, and the blue plane represents the mode boundary for the transporting mode. The red crosses denote an invariance failure, and the pink diamonds denote the attractor. Before modulation, there are trajectories prematurely exiting the mode; after modulation, all trajectories are bounded inside the mode.
  }
  \label{fig:3d_mod}
\end{figure}

We implemented the soup-scooping task on a Franka Emika robot arm. As depicted in Fig.~1, the task was to transport the soup (represented by the red beads) from one bowl to the other. Two demonstration trajectories were provided to the robot via kinesthetic teaching, from which we learned a DS to represent the desired evolution of the robot end-effector for each mode. The target velocity, $\dot{x}$, predicted by the DS was integrated to generate the target pose, which was then tracked by a Cartesian pose impedance controller. The robot state, $x$, was provided by the control interface. Sensor AP $r$ tracked the mode transition when the spoon made contact with the soup, and sensor AP $t$ tracked the mode transition when the spoon reached the region above the target bowl. $r$ and $t$ became true when the distance between the spoon and the centers of the soup bowl and the target bowl (respectively) were below a hand-tuned threshold. Sensor AP $s$ became true when red beads were detected either from a wrist camera via color detection or through real-time human annotation. We visualize the modulation of robot DS in three dimensions---$y$, $z$, and pitch---in Fig. \ref{fig:3d_mod}.

\begin{sidewaysfigure}
  \centering
  \includegraphics[width=\linewidth]{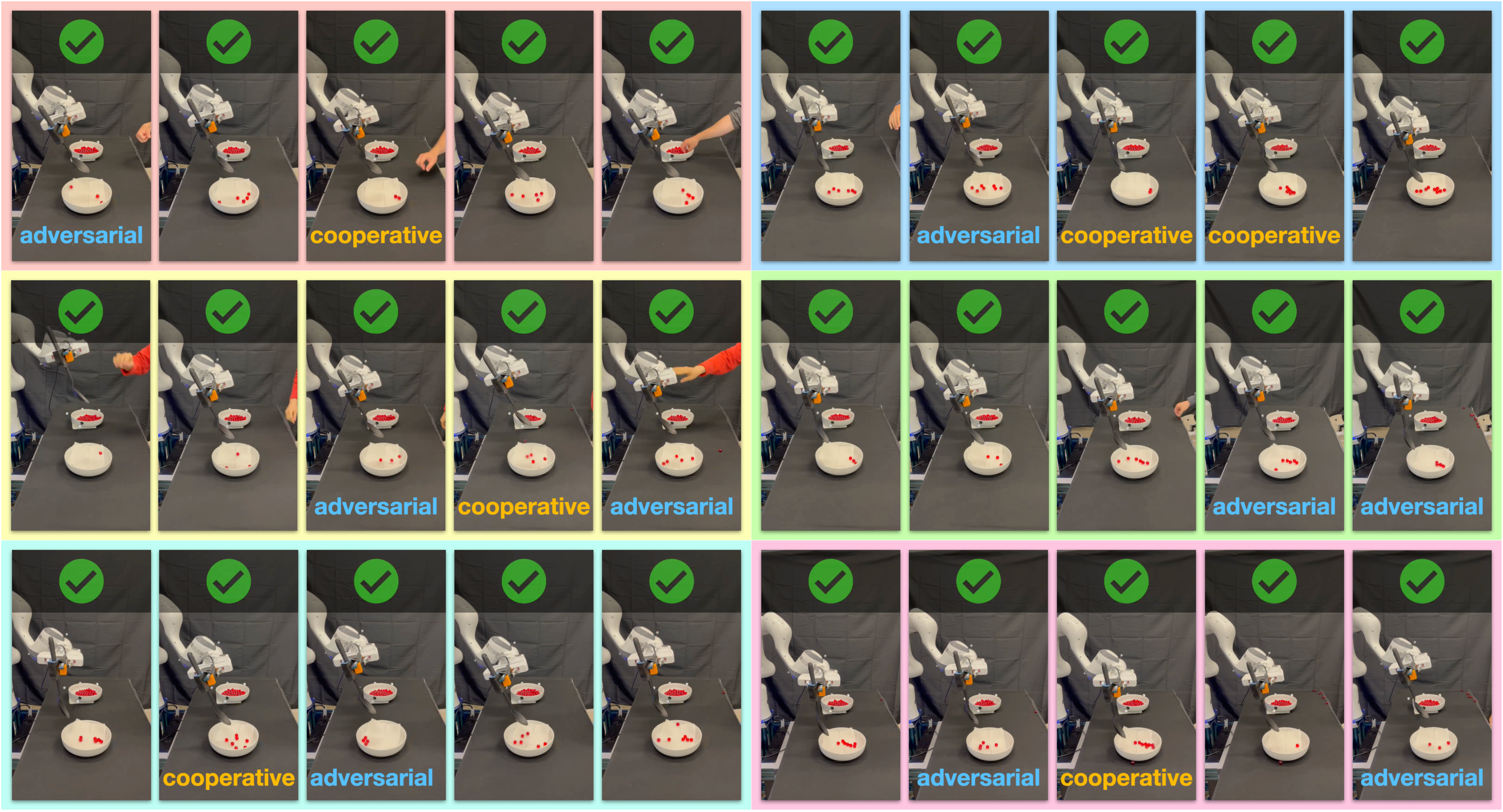}
  \caption{\footnotesize Ending snapshots (100\% success rate, see videos for action) of six randomly recruited human subjects performing unbiased perturbations in a total of 30 trials without cherry-picking. Common perturbation patterns (we annotate with the same colored text) emerge from different participants. Specifically, we see \textit{adversarial perturbations} where humans fight against the robot and \textit{cooperative perturbations} where humans help the robot to achieve the goal of transferring at least one bead from one bowl to the other.
  }
  \label{fig:unbiased}
\end{sidewaysfigure}

\textbf{Unbiased human perturbations} Since external perturbations are an integral part of our task complexity, we recruited six human subjects without prior knowledge of our LTL-DS system to perturb the robot scooping setup. Each subject is given five trials of perturbations. In total, we collected 30 trials as seen in Fig. \ref{fig:unbiased}, each of which is seen as an unbiased i.i.d. source of perturbations. On our project page, we show all 30 trials succeed eventually in videos. We did not cherry-pick these results, and the empirical 100\% success rate further corroborates our theoretic success guarantee. Interestingly, common perturbation patterns (as seen in the videos) emerge from different participants. Specifically, we see \textbf{adversarial perturbations} where humans fight against the robot and \textbf{cooperative perturbations} where humans help the robot to achieve the goal of transferring at least one bead from one bowl to the other. In the case of adversarial perturbations, DS reacts and LTL replans. In the case of collaborative perturbations, DS is compliant and allows humans to also guide the motion. In the case where humans are not perturbing yet the robot makes a mistake (e.g. during scooping), LTL replans the scooping DS until the robot enters the transferring mode successfully. The fact that we do not need to hard code different rules to handle invariance failures caused by either perturbations or the robot's own execution failures in the absence of perturbations highlights the strength of our LTL-powered sensor-based task reactivity. 

\section{Robot Experiment 2: Inspection Line} 

To further validate the LTL-DS approach we present a second experimental setup that emulates an inspection line, similar to the one used to validate the LPV-DS approach \cite{figueroa2018physically} -- which we refer to as the vanilla-DS and use to learn each of the mode motion policies.  In \cite{figueroa2018physically} this task was presented to validate the potential of the vanilla-DS approach to encode a highly-nonlinear trajectory going from \textcolor{red}{ (a) grasping region}, \textcolor{green}{ (b) passing through inspection entry,} (c) follow the inspection line and \textcolor{blue}{(d) finalizing at the release station} with a single DS. In this experiment we show that, even though it is impressive to encode all of these modes (and transitions) within a single continuous DS, if the sensor state or the LTL task-specification are not considered the vanilla-DS approach will fail to achieve the high-level goal of the task which is, to pass slide the object along the inspection line. To showcase this, in this work we focus only on \textcolor{green}{(b) $\rightarrow$}  \textcolor{blue}{(c) $\rightarrow$ (d)} with \textcolor{red}{ (a)} following a pre-defined motion and grasping policy for experimental completeness. 

\begin{figure}[!h]
  \centering
  \includegraphics[width=\linewidth]{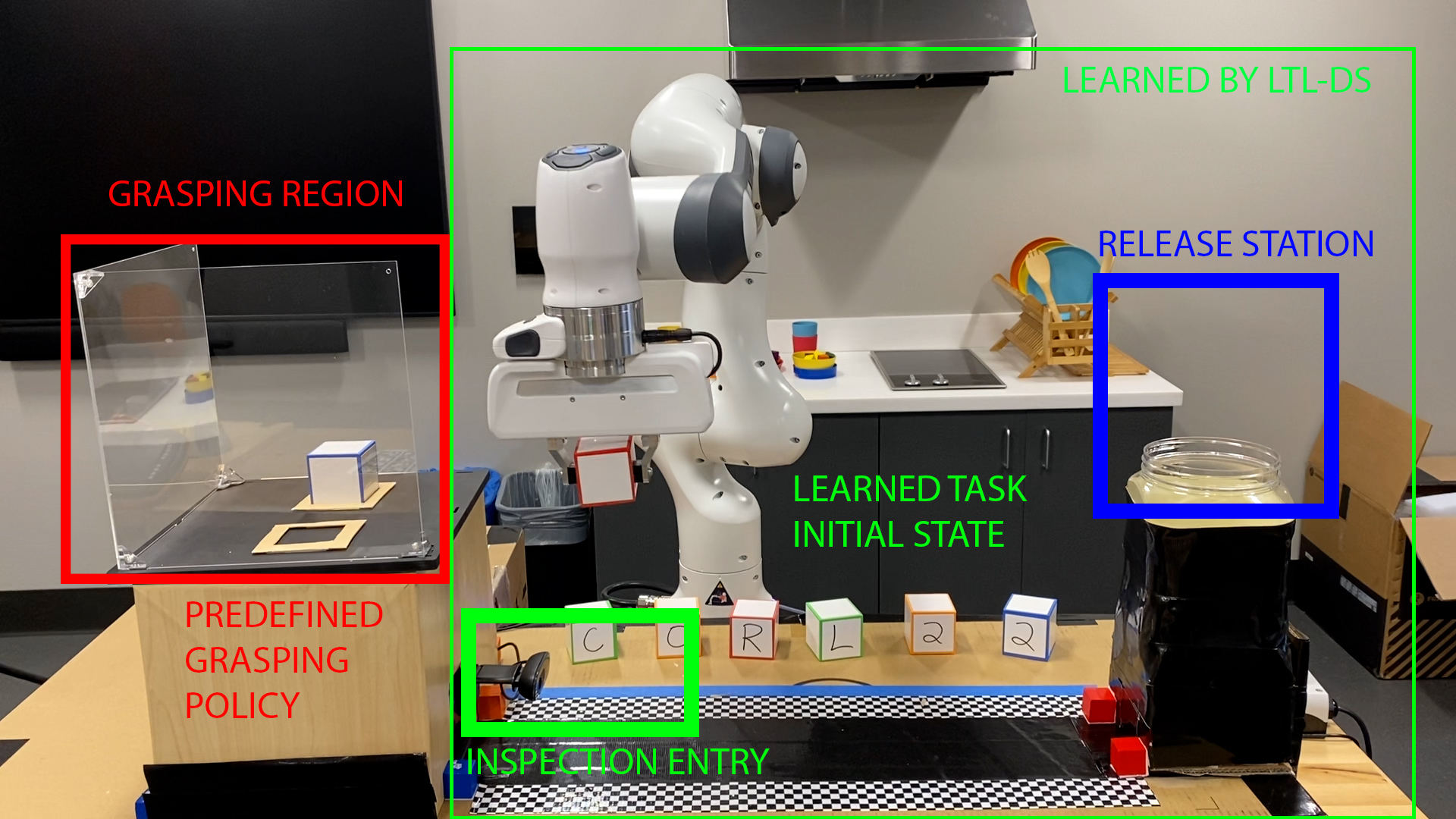}
  \caption{\footnotesize Robot Experiment 2: Inspection Line}
  \label{fig:inspection}
\end{figure}

\textbf{Inspection Task Details} \label{sec:inspection}
The video of this experiment can be found on our website.
\begin{itemize}[leftmargin=*]
    \item \textit{Sensor model:} We implement the \textit{sensor model} of the inspection task as an object detector on the inspection track and distances to attractors (defined from AP region-based segmentation described in new Appendix I). As we created a black background for the inspection task and the camera is fixed, with a simple blob detector we can detect if the robot is inside or outside of the inspection line. Hence, the \textit{sensor state} is a binary variable analogous to that of the scooping task.
    \item \textit{Task specification:} The proposed inspection task can be represented with 2 modes \textcolor{green}{(a) Go to inspection entry} $\rightarrow$ \textcolor{blue}{(b) follow inspection line and release}. The AP regions are the bounding boxes around the \textit{inspection entry} and \textit{release station} shown in Fig. \ref{fig:inspection} which correspond to the attractor regions for each mode. Mode (a) requires the robot to reach the mode attractor and detecting the presence of the cube once it has been reached. Mode (b) requires the cube to slide along the inspection track (reaching the end) and then lift the cube to drop it at the release station.
    
    \item \textit{Offline Learning:} We use two demonstrations of the inspection task, together with an LTL specification and run our offline learning algorithm used for the soup-scooping task (without any modifications), as shown in the supplementary video from \texttt{0:00-0:18s}. Without any adversarial perturbations or environmentally induced failures, the vanilla-DS approach is capable of accomplishing the defined inspection task without invariance failures as shown in \texttt{0:19-0:32s}.
    
    \item \textit{Invariance Failures of Vanilla-DS:} Even though the vanilla-DS approach is now used to learn a less complex trajectory (in terms of trajectory non-linearity), as we excluded the grasping region, we can see that it easily fails to achieve the inspection task when subject to large adversarial perturbations that lead the robot towards an \textit{out-of-distribution} state. This means that the robot was perturbed in such a way that it is now far from the region where the demonstrations were provided. Yet, it is robust to small adversarial perturbations that keep the robot \textit{in-distribution}, as shown in the supplementary video from \texttt{0:33-1:18min}. The latter is the strength of DS-based motion policies in general and these are the type of perturbations that are showcased in \cite{figueroa2018physically}. However, since the DS is only guaranteed to reach the target by solely imposing Lyapunov stability constraints it always reaches it after a large adversarial perturbation, with the caveat of not accomplishing the actual inspection task. Note that this limitation is not specific to the vanilla-DS approach \cite{figueroa2018physically}, it is a \textbf{general limitation} of goal-reaching LfD methods that only care about guaranteeing stability \textbf{at the motion-level} be it through Lyapunov or Contraction theory. \textit{Hence, by formulating the problem as TLI and introducing sensor states and LTL specification into the imitation policy we can achieve convergence at the motion-level and task-level.}\\
    
    
    \item \textit{Invariance Guarantee with LTL-DS:} As shown in the supplementary video from \texttt{1:19-1:43min} we collect a set of invariance failures to construct our mode boundary. Further, from \texttt{1:43-2:00min} we show the approximated mode boundary from 4 recorded failure states that approximate the vertical boundary and then from 10 recorded failure states which now approximate the horizontal boundary of the mode. The blue trajectories in those videos correspond to rollouts of the vanilla-DS learned from the demonstrations in that mode. 
    
    From \texttt{2:00-3:40min} we show two continuous runs of the inspection task, each performing two inspections. We stress test the learned boundary and LTL-DS approach by performing small and large adversarial perturbations. As shown in the video, when adversarial perturbations are small the DS motion policy is robust and still properly accomplishes the inspection task. When adversarial perturbations are large enough to push the robot outside of the learned boundary, the LTL-DS brings the robot back to the inspection entry mode and tries the inspection line again and again and again until the inspection task is achieved as defined by the LTL specification - \textbf{guaranteeing task completion.} \\
\end{itemize}

\textbf{Comment on Task Definition:} In order to learn and encode the entire task (from grasp to release) with LTL-DS we would need to include a grasping controller within our imitation policy. It is possible to extend the LTL-DS approach to consider grasping within the imitation policy, yet due to time limitations we focus solely on the parts of the task that can be learned by the current policy -- that requires only controlling for the motion of the end-effector. We are concurrently working on developing an approach to learn a grasping policy entirely through imitation, which to the best of our knowledge does not exist within the problem domains we target. In a near future, we plan to integrate these works in order to allow LTL-DS to solve problems that include actuating grippers in such a feedback control framework. Note that, the vanilla-DS approach does not consider the grasping problem either and the experimental setup presented in \cite{figueroa2018physically} uses a simple open-loop gripper controller that is triggered when the DS reaches the attractor, such triggering is hand-coded and not learned either in their setup.

\section{Robot Experiment 3: Color Tracing} \label{sec:color}

\begin{figure}[!h]
  \centering
  \includegraphics[width=1\linewidth]{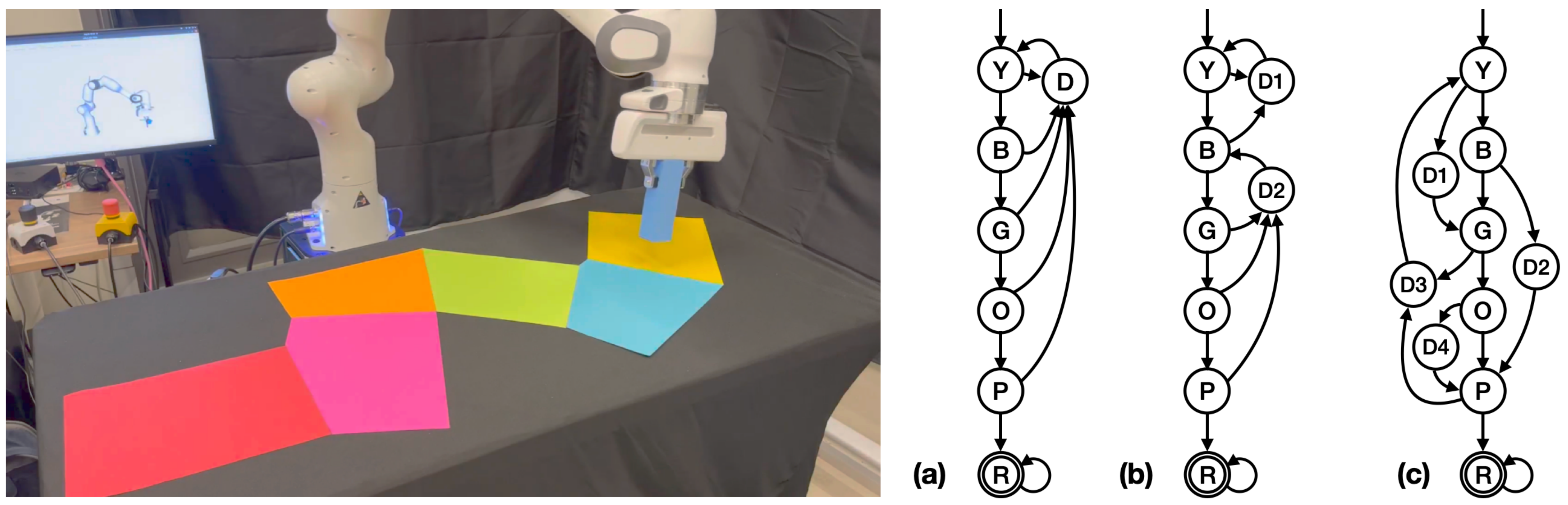}
  \caption{\footnotesize The goal is for the end-effector to move through (while staying within) (Y)ellow, (B)lue, (G)reen, (O)range, (P)ink, and eventually reach (R)ed. If a sensor detects the end-effector is perturbed into the (D)ark region, the system needs to replan according to a given task automaton such as (a), (b), (c). Note D1, D2, D3, D4 refer to different modes (entering the dark region from different colors) that appear visually the same.}
  \label{fig:color}
\end{figure}

This experiment demonstrates LTL-DS' handling of long-horizon multi-step tasks with non-trivial task structures. Given a single demonstration of kinesthetically teaching the robot end-effector to move through the colored tiles, the system learns a DS for each of the colored mode. The learned DS can then be flexibly recombined according to different LTL-equivalent task automatons to react differently given invariance failures. Specifically, we show in the videos on our website  three different replanning: (a) mode exit at any colored tile transitions to re-entry at the yellow tile; (b) mode exit at any colored tile after the blue tile transitions to re-entry at the blue tile; and (c) mode exit at the yellow tile transitions to the blue tile, while mode exit at the blue tile transitions to the pink tile.




\clearpage

\end{document}